\RequirePackage{fix-cm}
\documentclass[smallextended]{svjour3}       
\smartqed  
\usepackage{graphicx}
\graphicspath{{EB-GLS_paper_resource/}}
\usepackage{array}
\usepackage{algorithm}
\usepackage{algorithmic}
\usepackage{multirow}
\usepackage{subfigure}
\usepackage{amsmath,amsfonts,amssymb}
\usepackage{appendix}

\usepackage{color}

\begin{document}

\title{EB-GLS: An Improved Guided Local Search Based on the Big Valley Structure
}


\author{Jialong~Shi \and Qingfu~Zhang \and Edward~Tsang}


\institute{Jialong Shi \at
              Department of Computer Science, City University of Hong Kong, Hong Kong\\
              \email{jlshi2-c@my.cityu.edu.hk}           
           \and
           Qingfu Zhang \at
              Department of Computer Science, City University of Hong Kong, Hong Kong\\
           \and
           Edward Tsang \at
              Centre for Computational Finance and Economic Agents, School of Computer Science and Electronic Engineering, University of Essex, Colchester, UK\\
}

\date{Received: date / Accepted: date}

\maketitle

\begin{abstract}
  Local search is a basic building block in memetic algorithms. Guided Local Search (GLS) can improve the efficiency of local search. By changing the guide function, GLS guides a local search to escape from locally optimal solutions and find better solutions. The key component of GLS is its penalizing mechanism which determines which feature is selected to penalize when the search is trapped in a locally optimal solution. The original GLS penalizing mechanism only makes use of the cost and the current penalty value of each feature. It is well known that many combinatorial optimization problems have a big valley structure, i.e., the better a solution is, the more the chance it is closer to a globally optimal solution. This paper proposes to use big valley structure assumption to improve the GLS penalizing mechanism. An improved GLS algorithm called Elite Biased GLS (EB-GLS) is proposed. EB-GLS records and maintains an elite solution as an estimate of the globally optimal solutions, and reduces the chance of penalizing the features in this solution.  We have systematically tested the proposed algorithm on the symmetric traveling salesman problem. Experimental results show that EB-GLS is significantly better than GLS.
\end{abstract}

\section{Introduction}
Memetic algorithms~\cite{hasan2009memetic} use Local Search (LS) to improve their efficiency. Usually the solutions LS stops at are locally optimal solutions. Guided Local Search (GLS)~\cite{voudouris2010guided} is a strategy that can improve the efficiency of LS. By changing the guide function of LS, GLS guides a LS procedure escape from locally optimal solutions and find better solutions. In this paper, we propose and study some improvements of the basic GLS strategy. The improved search strategy can be used by the memetic algorithm in the future. Our improvements are based on the big valley structure assumption in combinatorial optimization problems.

To use GLS, one has to first define a set of features which a candidate solution may exhibit. When the LS procedure is trapped in a local optimum, some selected unfavorable features are penalized. The objective function is augmented by the accumulated penalties and then used to guide the further search to move out of the attraction region of this local optimum. How to select features to penalize is a major issue in GLS. The penalizing mechanism proposed in~\cite{voudouris2010guided} only considers the cost and the current penalty value of each feature. Based on the ``big valley'' structure assumption in combinatorial optimization~\cite{boese1995cost}, this paper proposes to estimate how likely each feature appears in a globally optimal solution from the best solutions found so far, and then use such information to improve the penalizing mechanism of GLS. More specifically, our proposed GLS method, called Elite Biased Guided Local Search (EB-GLS), record a best solution found so far during the search. Features in this solution are assumed to have a high chance to be good, and their probabilities to be penalized are reduced by using a simple method. The symmetric Traveling Salesman Problem (TSP) is used as a test suite to study the proposed EB-GLS in this paper. Our aim is not to develop the best algorithm for the TSP, but to illustrate that our modification can significantly improve the performance of GLS.

This paper is structured as follows. Section \ref{sec:GLS} presents the GLS procedure for the TSP. Section \ref{sec:liter} introduces the recent works on designing an improved version of GLS. Section \ref{sec:big_valley} discusses the big valley structure in the symmetric TSP. Section \ref{sec:EBGLS} presents the EB-GLS procedure for the symmetric TSP. To investigate whether or not EB-GLS achieves its design goal, Sections \ref{sec:EBGLS_behavior} and \ref{sec:perform_compare} compares EB-GLS and GLS experimentally. Our experiments are conducted on symmetric TSP instances from the TSPLIB and randomly generated symmetric TSP instances. Section \ref{sec:conclusion} concludes the paper.

\section{Guided Local Search} \label{sec:GLS}
\subsection{Traveling Salesman Problem}
Let $G=(V, E)$ be a fully connected graph where $V$ is its node set and $E$ the edge set, and let $c_e>0$ be the cost of $e \in E$. A tour $s$ in $G$ is a cycle passing through every node in $V$ exactly once and its cost is defined as:
\begin{equation}
g(s)=\sum_{e \in s}c_e.
\end{equation}
A node in $G$ can be interpreted as a city and $c_e$ as the travel cost from the source node of edge $e$ to its destination node. $g(\cdot)$ is the objective function. The goal of the Traveling Salesman Problem (TSP) is to find a tour with the smallest $g$ value. The TSP is one of the most widely used NP-hard test problems in the area of heuristics. There are many different TSP variants. This paper considers the symmetric TSP, where $G$ is undirected, i.e., the cost of travel from node $A$ to node $B$ is the same as that from $B$ to $A$. We choose test instances from the best known TSPLIB~\cite{reinelt1991tsplib} in our experimental studies. The instances in TSPLIB have known globally optimal costs. The dimensionality of a instance in TSPLIB is reflected by its name. For example, the dimensionality of the instances att532 is 532. We denote the set of all the tours in $G$ as $S$, which is the solution space of the TSP.

\subsection{Local Search}
LS is a basic search and optimization technique. It can be used as an improved technology for many existing algorithms~\cite{jadon2015accelerating,marinaki2015hybridization}. For example, LS has been successfully incorporated in the memetic algorithm to exploit the problem knowledge. LS defines a neighborhood for every candidate solution in the search space. It maintains one candidate solution and iteratively improves it. It searches the neighborhood of the current solution and moves to a neighboring solution which has a better guide function value. Most LS algorithms use the objective function of the problem in question as their guide function. LS stops and outputs the current solution as its final solution to the problem when all the neighbors are not better than the current solution according to the guide function. Since the neighborhood size is limited, LS usually stops at solutions that are not worse than their neighbors but not necessarily all other solutions in the search space, i.e. the locally optimal solutions.

Commonly-used LS heuristics for the TSP include 2-Opt heuristic, 3-Opt heuristic and Lin-Kernighan (LK) heuristic. All these algorithms are based on edge exchange. 2-Opt heuristic replaces two edges of the current solution by two other edges to obtain a neighboring solution. In 3-Opt heuristic, the number of edges to change is 3. In LK heuristic, the number of edges to change is variable.

\subsection{Procedure of Guided Local Search}
GLS is a simple penalty-based approach for helping a LS procedure to escape from local optima by dynamically adjusting its guide function. To use GLS, one needs to define a set of (solution) features for the given problem. For example, in the TSP, the features can be defined as the edges between nodes. Given a candidate solution $s$ and a feature $i$, function $I_i(s)$ is an indicator function of whether solution $s$ exhibits feature $i$:
\begin{equation}
I_i(s)= \left\{
    \begin{array}{rl}
        1 &\mbox{ if feature $i$ is in $s$}, \\
        0 &\mbox{ otherwise.}
    \end{array} \right.
\end{equation}

In GLS, each feature has a cost and a penalty. The cost is related to the objective function $g$. For example, in the TSP the cost of a feature is the cost of the corresponding edge. The penalties of all features are initialized to be zero at the beginning. Unlike other LS heuristics, GLS does not use the original objective function $g(\cdot)$, but function $h(\cdot)$ as its guide function for its LS procedure. Function $h(\cdot)$ is defined as:
\begin{equation}\label{eq:h_function}
h(s)=g(s)+ \lambda\sum p_iI_i(s),
\end{equation}
where $\lambda$ is the parameter that controls the penalizing strength, $p_i$ is the current penalty value of feature $i$. We call $h(\cdot)$ the augmented objective function.

GLS starts from an initial solution and executes a LS at each iteration using $h(\cdot)$ as its guide function. Once the LS stops at a local optimum $s_*$. GLS adjusts $h(\cdot)$ by increasing the penalties of one or more selected features in $s_*$. To do so, GLS defines the utility of each feature $i$, $util_i$ as
\begin{equation}\label{eq:def_of_util}
util_i(s_*)=I_i(s_*)\cdot \frac{c_i}{1+p_i},
\end{equation}
where $c_i$ is the cost of feature $i$. GLS selects the features with the highest utility value, and increases their penalties by 1. Then a new iteration starts from $s_*$. The pseudocode of GLS is shown in Algorithm \ref{alg:GLS}. Inputs are the objective function $g$, the GLS parameter $\lambda$, the feature set $M$ and the cost of each feature $i \in M$.

\begin{algorithm}
    \begin{algorithmic}[1]
        \STATE \textbf{input:} $g,\lambda,M, c$
            \STATE $j \gets 0$
            \STATE $s_0 \gets $ random or heuristically generated solution.
            \FOR {$i=1\to |M|$}
                \STATE $p_i \gets 0$
            \ENDFOR
            \WHILE {!StoppingCriterion}
                \STATE $h\gets g+\lambda \sum p_iI_i$
                \STATE $s_{j+1} \gets$ LocalSearch($s_j,h$) 
                \FOR {$i = 1\to |M|$}
                    \STATE $util_i \gets I_i(s_{j+1})\cdot c_i/(1+p_i)$
                \ENDFOR
                \FOR {each $i$ such that $util_i$ is maximum}
                    \STATE $p_i\gets p_i + 1$
                \ENDFOR
                \STATE $j\gets j+1$
            \ENDWHILE
        \STATE $s_*\gets$ the best solution found with respect to $g$
        \RETURN{$s_*$}
    \end{algorithmic}
\caption{Guided Local Search}
\label{alg:GLS}
\end{algorithm}

GLS conducts LS based on the augmented objective function $h(\cdot)$, which is different from the original objective function $g(\cdot)$. Hence GLS has to record the best solution found so far with regard to $g$. After each move of LS, GLS will check whether the $g$ value of the new solution is lower than that of the recorded best solution, if so, the recorded best solution will be updated.

GLS has been successfully applied to the TSP~\cite{voudouris1999guided}. To apply GLS to the TSP, we set edges as features and the costs of edges as the costs of features. Therefore, $M=E$. In this paper, we use 2-Opt as the LS heuristic of GLS on the TSP, because according to \cite{voudouris1999guided} GLS performs better with 2-Opt, especially when it is combined with the Fast Local Search (FLS) strategy~\cite{bentley1992fast,voudouris1999guided}. In addition, using the 2-Opt LS heuristic makes our proposed algorithm easy to be implemented by other researchers.

\subsection{Remarks}
\begin{figure}
  \centering
  \subfigure[ILS]{
  \label{fig:ILS_escape_LO}
  \includegraphics[width=0.9\textwidth]{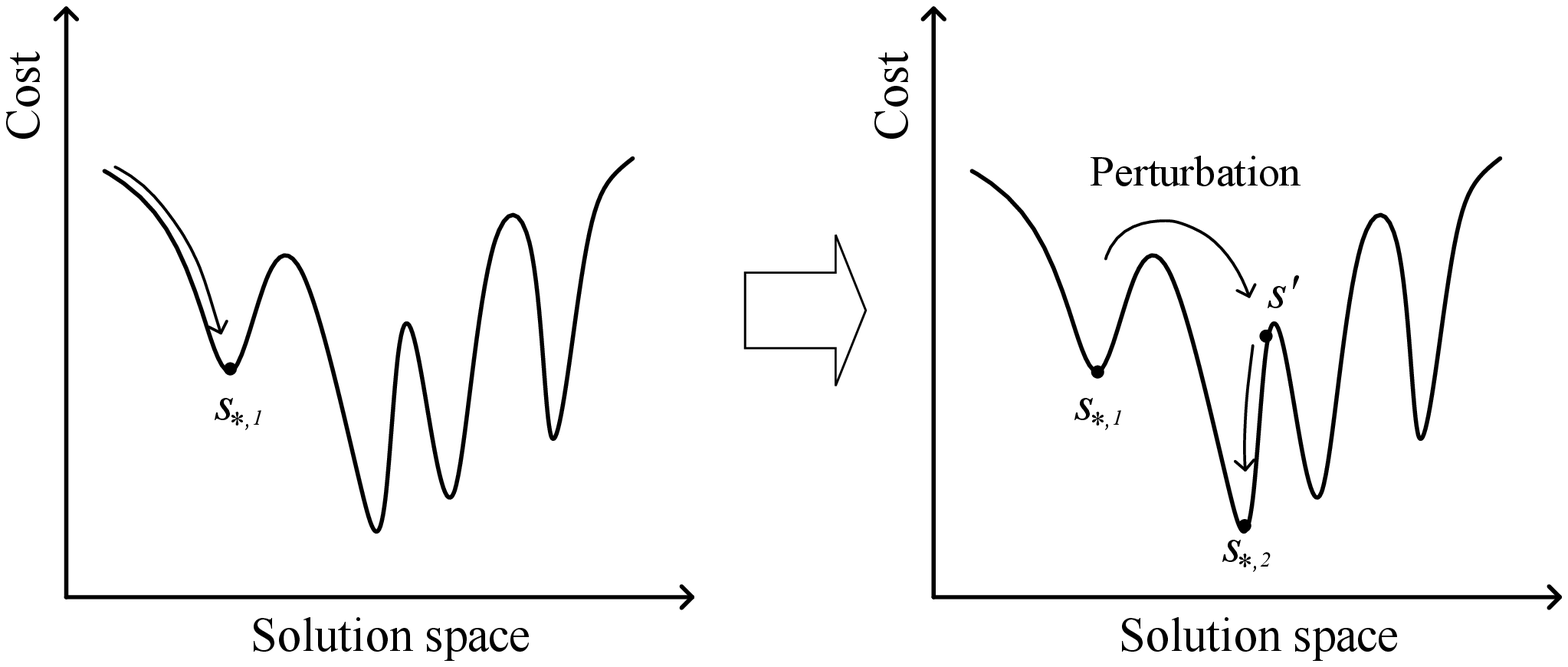}}
  \subfigure[GLS]{
  \label{fig:GLS_escape_LO}
  \includegraphics[width=0.9\textwidth]{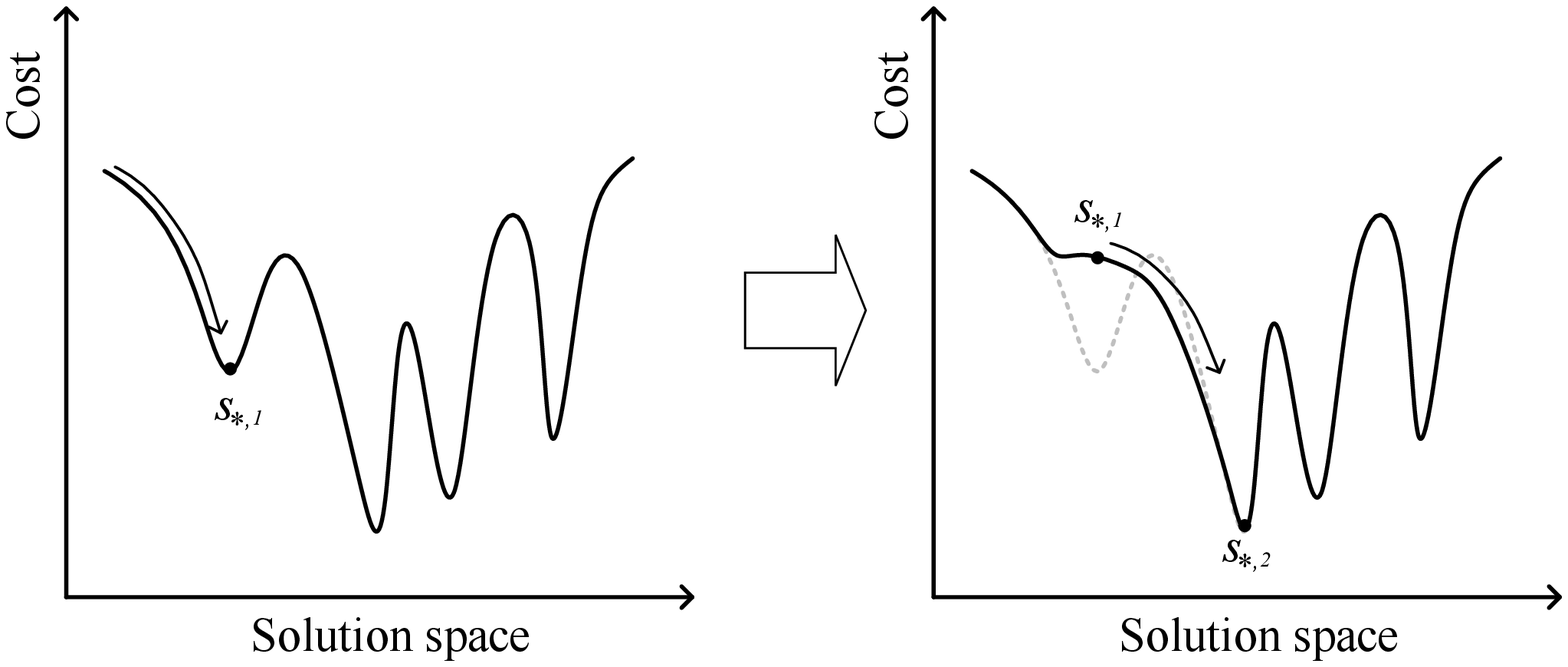}}
  \caption{The comparison between ILS's procedure and GLS's procedure}\label{fig:GLSvILS} 
\end{figure}

At each iteration, GLS performs a LS procedure and tries to escape from the encountered local optimum, which is similar to a well-known metaheuristic, Iterated Local Search (ILS)~\cite{lourencco2010iterated}. ILS tries to escape from the current local optimum by perturbation, as illustrated in Fig. \ref{fig:ILS_escape_LO}. GLS tries to escape from the current local optimum by increasing penalty on it. The increased penalty changes the guide function of GLS, which can be seen as ``lifting'' the local optimum, as illustrate in Fig. \ref{fig:GLS_escape_LO}.
The comparison results in \cite{voudouris1999guided} show that GLS performs better than ILS under the same computation cost. This means that in GLS the changed guide function improves the efficiency of the LS procedure. This is due to the well-designed penalizing mechanism of GLS. In (\ref{eq:def_of_util}), a feature with low cost $c$ and the high current penalty value $p$ are regarded as good. Good features in the current solutions will have little chance to be penalized. Here we analyse the design of (\ref{eq:def_of_util}) based on the knowledge (information) it learns during the search. The basic consideration in (\ref{eq:def_of_util}) is:
\begin{itemize}
  \item \textbf{Use of a priori problem specific knowledge}:  The goal of the TSP is to minimize the total cost of the tour.  It is very natural to encourage to use low-cost features (i.e., edges in the TSP).
  \item  \textbf{Use of online knowledge learned from the search}: If a feature exhibited in the current solution has a high penalty value,  it implies that this feature has been penalized for many times. Such a feature is likely to be part of a good solution since it has exhibited in many locally optimal solutions of $h$.
\end{itemize}
The online knowledge used in GLS is the current penalty value of features. We believe that other forms of online knowledge can be exploited to improve estimation of goodness of a feature. This paper represents our effort along this line. Our work in this paper is based on the so called big valley structure hypothesis in combinatorial optimization.

\section{Literature Review}\label{sec:liter}
Besides the TSP, GLS has been successfully applied to many other optimization problems~\cite{voudouris2010guided}. In addition, a number of variants and hybrid algorithms based on GLS have been proposed.

Guided Genetic Algorithm (GGA)~\cite{lau2001guided} is a hybrid algorithm of GLS and Genetic Algorithm (GA). In GGA, when GA does not find an improved solution for a number of generations, the penalizing mechanism of GLS will be executed to change the guide function of GA. GGA is different from the Stepwise Adaptation of Weights (SAW) mechanism~\cite{eiben1998graph} which also changes the guide function of EA. When SAW increases penalties (weights) on the best individual, it increases penalties on all the violated constrains. When GGA increases penalties on the best individual, it only increases penalties on the features that have the largest $util$ value, and $util_i = c_i/(1+p_i)$ is related to the historical penalty $p_i$. Hence, compared to SAW, the penalizing mechanism of GGA (which comes from GLS) considers the historical penalties to avoid penalizing a feature too many times. In the works of Mills et al.~\cite{mills2003applying}, two extended versions of GLS are proposed. The first one involves the aspiration criteria, which means ignoring the penalties if a move can produce a new best solution. The second extended GLS allows random move from the neighbourhood when certain conditions are satisfied. Basharu et al.~\cite{basharu2005distributed} proposed a modified GLS to solve the distributed constraint satisfaction problems, which is called Distributed GLS (Dis-GLS). In Dis-GLS, additional heuristics are introduced so that it can handle distributed scenarios. Tao and Haubrich~\cite{tao2005hybrid} proposed a hybrid algorithm of GLS and Large Neighborhood Search (LNS) for the planning of medium-voltage power distribution systems, in which GLS and LNS are performed in different phases. In the GLS implementation proposed by Zhong and Cole~\cite{zhong2005vehicle}, the features whose utilities are larger than a certain value are penalized and the penalties have an upper bound. In the work of Mester and Braysy~\cite{mester2005active}, a hybrid algorithm of GLS and Evolution Strategies (ES) called Active Guided Evolution Strategies (AGES) is proposed. In AGES, when no improvement have been made for a user-defined number of iterations, the penalties are cleared and the search is restarted from the historical best solution. Guided Tabu Search (GTS)~\cite{tarantilis2008hybrid} is a hybrid algorithm of GLS, Variable Neighborhood Search (VNS) and Tabu Search (TS). In GTS, the guide function of TS is the objective function augmented by the penalty terms, which is same as that in GLS. In addition, GTS uses a new utility function $util_i = (c_i/avg_i)/(1+p_i)$, in which the term $avg_i$ can be considered the average cost of the other features related to feature $i$. The penalties on features are temporary and will be cleared after a certain tenure. In a variant of GTS~\cite{zachariadis2009guided}, the costs of the features with the largest utility will be multiplied by 2, and restore to the original values after a given tenure. In the GLS implementation developed by Vansteenwegen et al.~\cite{vansteenwegen2009guided}, when GLS is in a local optimum, it penalizes the included features with the highest ``disutility'' and rewards the non-included features with the highest utility. Here the principle of calculating utility and disutility is same as that of GLS.

GLS also has been applied to multi-objective optimization problems. In the related works, a multi-objective version of GLS called Guided Pareto Local Search (GPLS)~\cite{alsheddy2011empowerment2} and a hybrid of GLS with MOEA/D~\cite{alhindi2013moea} are proposed. However, in this paper we only discuss the possible improvement of GLS on single-objective optimization problems.

In the aforementioned variants and hybridizations of GLS, not much effort has been done to enhance the penalizing mechanism of GLS. In GTS, the penalizing mechanism considers the relative magnitude of feature cost through dividing the term $avg_i$. But there is no systemic experiment to prove that this modification in GTS can bring significant performance improvement. In \cite{vansteenwegen2009guided}, the features not included in current local optimum undergo an inverse version of the penalizing mechanism, but no systemic experiment study is done too. In this paper, we propose an enhanced penalizing mechanism of GLS by exploiting the big valley assumption. We also conduct systemic experiments to show that the modified penalizing mechanism can improve the performance of GLS on the symmetric TSP.

\section{Big Valley Structure} \label{sec:big_valley}
Boese~\cite{boese1995cost} presents a scatter plot of different local optima of the TSP instance att532 to illustrate its landscape. The following bond distance between two solutions $s_1$ and $s_2$:
\begin{equation}\label{eq:tsp_dist}
  \mbox{distance}(s_1,s_2) = |\{e\in E|e\in s_1 \land e\notin s_2\}|
\end{equation}
is used in his studies. He observed that there is a strong correlation between the distance of a solution to a global optimum and its cost. He call this phenomenon the big valley structure. Kauffman~\cite{kauffman1993origins} observed the similar phenomenon in the NK landscapes with small K values, which is called ``Massif Central''. Jones and Forrest~\cite{jones1995evolutionary,jones1995fitness} also studied the relationship between the solution cost (which they call fitness) and the distance to the global optimum. They introduced fitness distance correlation (FDC) to measure the correlation between the solution quality and the distance to the nearest global optimum:
\begin{equation}\label{eq:FDC}
  \mbox{FDC}(g,d_{opt}) = \frac{\mbox{cov}(g,d_{opt})}{\sigma(g)\sigma(d_{opt})},
\end{equation}
where $g$ is the objective function value, i.e. the cost of solution, $d_{opt}$ is the distance to the nearest global optimum, $\mbox{cov}(\cdot)$ denotes the covariance and $\sigma(\cdot)$ denotes the standard deviation. They suggested that FDC should be explored in algorithm design. Merz and Freisleben~\cite{merz2001memetic} conducted fitness landscape analysis on several TSP instances and found out that the fitness and the distance to the optimum are highly correlated for most instances. The big valley structure has also been reported in other combinatorial optimization problems, such as the maximum satisfiability~\cite{zhang2004configuration}, the unconstrained binary quadratic programming~\cite{merz2004memetic}, the quadratic assignment problem~\cite{merz2000fitness}, the flowshop scheduling~\cite{reeves1999landscapes}, and others.

Although the big valley structure has been reported in many publications for a wide range of problems, however there is no uniform and strict definition for the big valley structure. In some publications, the big valley structure means that the global optima and local optima are clustered in a small region of the solution space. Some other publications mean high correlation between the cost and the distance to the nearest global optimum.

\begin{figure}
  \centering
  \includegraphics[width=0.8\textwidth]{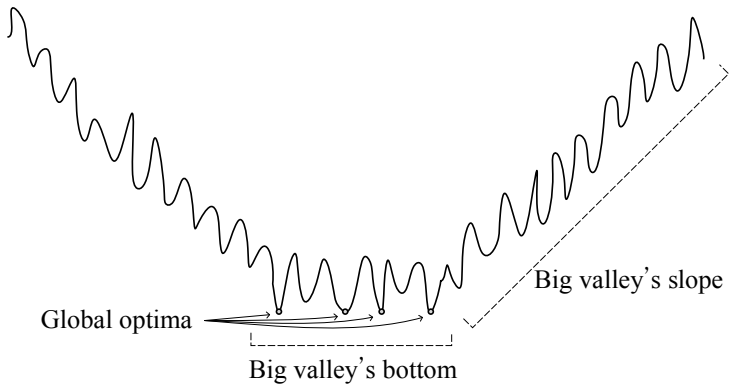}\\
  \caption{Depiction of the big valley structure defined in this paper }\label{fig:big_valley}
\end{figure}

In this paper, we say that a TSP instance has a big valley structure when the following two requirements are met.
\begin{itemize}
  \item If there are multiple globally optimum solutions, the mean bond distance between two global optima is significantly smaller than the mean distance between two randomly selected solutions, i.e. $N/2$. In other words, the global optima are located in a small region of the solution space, which can be seen as the ``bottom'' of the big valley, and
  \item There is a strong correlation between the cost of a solution and its distance to the nearest global optimum, i.e., the FDC value is relatively large.
\end{itemize}
Fig. \ref{fig:big_valley} illustrates the big valley structure we mean in this paper.

\begin{figure}
  \centering
  \includegraphics[width=0.8\textwidth]{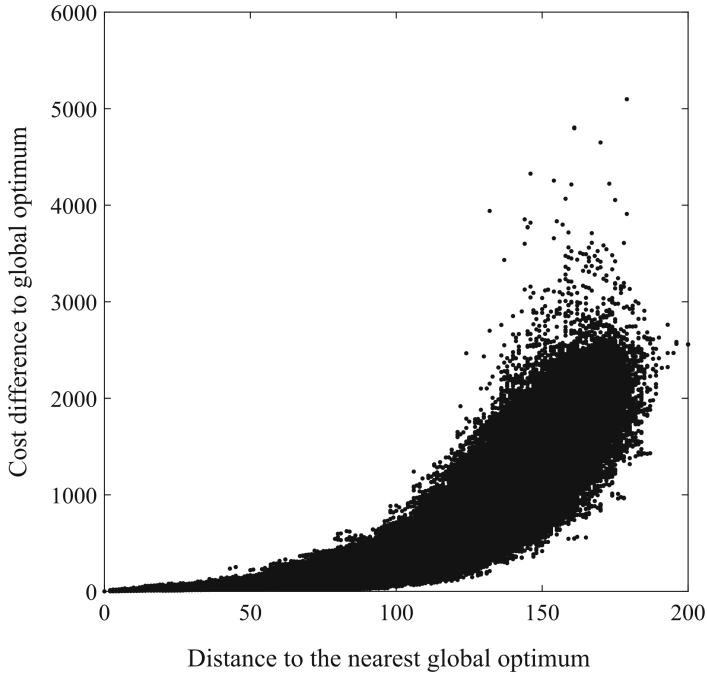}\\
  \caption{The recorded solutions during \text{1,000} runs of GLS and \text{1,000} runs of EB-GLS on att532. The cost difference to the globally optimal cost (vertical axis) is plotted against the distance to the nearest globally optimal solution (horizontal axis). There is a clear correlation between the cost and the distance to the nearest global optimum.}\label{fig:CvD_att532} 
\end{figure}

Let's consider the TSP instance att532 with 532 nodes. According to \cite{hains2011revisiting,ochoa2016deconstructing}, It has two different global optima, and the bond distance between these two global optima is 2. Thus it meets the first requirement. To study if it meets the second requirement, we have executed 1,000 times of GLS and 1,000 times of EB-GLS (the GLS variant proposed in this paper) on att532 independently from randomly generated solutions. During each run, when the current best solution changes, the new best solution is recorded. In total, \text{292,714} solutions have been recorded. Fig. \ref{fig:CvD_att532} is the scatter plot of these solutions. The FDC value of the solutions shown in Fig. \ref{fig:CvD_att532} is 0.83. It is clear that it meets the second requirement. Therefore, the instance att532 has the big valley structure, which is consistent with the statement made by Boese~\cite{boese1995cost}.

However, the big valley structure does not hold in all the TSP instance. We have conducted the same experiments on nine other instances from the TSPLIB, and found that eight instances have the big valley structure while one instance does not. The experiment results on the eight instances can be found in the Appendix. Here we focus on the instance that does not have the big valley structure, u2319. On u2319, \text{2,000} runs of GLS/EB-GLS found total \text{2,000} different global optima, which means that u2319 have at least \text{2,000} global optima. The mean distance between these global optima is 884, in which the minimum distance is 777 and the maximum distance is 992. Hence the mean distance between the global optima is not significantly smaller than the mean distance between two randomly selected solutions in u2319, i.e. 2319/2 = 1160. This means that the global optima are widely scattered in the solution space. Thus we claim that u2319 does not satisfy the first requirement of the big valley structure.

In the following, we propose an enhanced GLS algorithm which exploits the big valley structure assumption. We will show that the proposed algorithm performs significantly better than GLS on att532 (the instance that has the big valley structure). Meanwhile its performance does not drastically deteriorate on u2319 (the instance that does not have the big valley structure).

\section{Elite Biased Guided Local Search} \label{sec:EBGLS}
\subsection{Procedure}
Our improved GLS algorithm called Elite Biased Guided Local Search (EB-GLS) maintains an elite solution $s_e$ (i.e. a high-quality solution) as an estimate of the global optimum, and uses a different $util$ formula. The new $util$ formula is:
\begin{equation}\label{eq:def_of_util_EBGLS}
util_i(s_*)= \left \{
\begin{array}{ll}
I_i(s_*)\cdot \displaystyle{\frac{c_i}{1+p_i}}, &\mbox{if } s_e \mbox{ includes feature } i;\\
I_i(s_*)\cdot \displaystyle{\frac{c_i}{1+p_i}} \cdot w, &\mbox{otherwise},\\
\end{array}
\right.
\end{equation}
where $w$ is a predefined constant larger than 1. Since $w>1$, the features in $s_e$ will have relatively small $util$ values, hence the penalties imposed on the features in $s_e$ are reduced. EB-GLS directly use the best solution found so far as $s_e$, so that no additional effort is introduced. Algorithm \ref{alg:EBGLS} is the pseudocode of EB-GLS. Compared to GLS (Algorithm \ref{alg:GLS}), EB-GLS has an extra input parameter: $w$.

\begin{algorithm}
    \begin{algorithmic}
        \STATE \textbf{input:} $g,\lambda,M,c,w$
            \STATE $j \gets 0$
            \STATE $s_0 \gets $ random or heuristically generated solution in $S$
            \FOR {$i=1\to |M|$}
                \STATE $p_i \gets 0$
            \ENDFOR
            \WHILE {!StoppingCriterion}
                \STATE $h\gets g+\lambda \sum p_iI_i$
                \STATE $s_{j+1} \gets$ LocalSearch($s_j,h$) 
                \STATE $s_e \gets$ historical best solution with respect to $g$
                \FOR {$i = 1\to |M|$}
                    \IF {$s_e$ includes feature $i$}
                        \STATE $util_i \gets I_i(s_{j+1})\cdot \frac{c_i}{(1+p_i)}$
                    \ELSE
                        \STATE $util_i \gets I_i(s_{j+1})\cdot \frac{c_i}{(1+p_i)} \cdot w$ \qquad /* $w>1$ */
                    \ENDIF
                \ENDFOR
                \FOR {each $i$ such that $util_i$ is maximum}
                    \STATE $p_i\gets p_i + 1$
                \ENDFOR
                \STATE $j\gets j+1$
            \ENDWHILE
        \STATE $s_*\gets$ historical best solution with respect to $g$
        \RETURN{$s_*$}
    \end{algorithmic}
\caption{Elite Biased Guided Local Search}
\label{alg:EBGLS}
\end{algorithm}

The efficiency and effectiveness of a search algorithm depend on how search effort is allocated in the search space, and balancing between exploitation and exploration. The aim of EB-GLS is to allocate more search effort on the search regions near to $s_e$ by reducing the penalties on $s_e$. According to the big valley structure hypothesis, those regions are more likely to contain high-quality solutions, even the global optima. Meanwhile EB-GLS keeps updating $s_e$ by the newly found better solutions to explore more promising regions and prevent the search process from stalling.


\subsection{Some Implementation Notes}

\subsubsection{The update of $s_e$}

At the early search stage, a best solution found so far may be not of high quality and thus can be far away from the global optima. On the other hand, too frequent update of $s_e$ in EB-GLS is not necessary and may involve extra computation load. Based on these considerations, in our experimental studies, suppose the total runtime is $T$, we first run GLS for $10/T$ and then change to EB-GLS. We update $s_e$ once every 100 executions of the LS procedure.

\subsubsection{The values of $\lambda$ and $w$}
Following \cite{voudouris1999guided}, we set
\begin{equation}\label{eq:lambda}
\lambda = 0.3 \cdot \frac{g(\mbox{first local optimum})}{N},
\end{equation}
where $g(\mbox{first local optimum})$ is the cost of the first local minimum. As to $w$, our pilot experiments show that EB-GLS is not very sensitive to $w$. We set $w=2$ in this paper.

\section{Behavior of Elite Biased Guided Local Search} \label{sec:EBGLS_behavior}
\begin{figure}
  \centering
  \includegraphics[width=\textwidth]{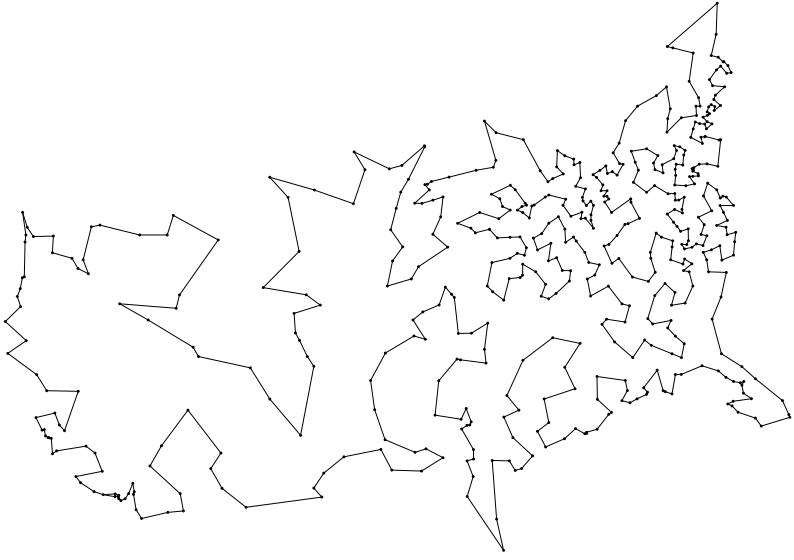}\\
  \caption{Edges that belong to the two global optima of att532}\label{fig:att532GOE} 
\end{figure}

EB-GLS aims at reducing penalties on the features in the global optima by exploiting the big valley structure. To investigate whether EB-GLS achieves this design goal, we conduct an experiment which records the behaviors of EB-GLS and GLS during the search. We select two instances, att532 and u2319, as the test instances. According to Section \ref{sec:big_valley}, att532 has a typical big valley structure, meanwhile u2319 does not have the big valley structure. In our experiment, we execute thousands of runs of GLS and EB-GLS on att532 and u2319. Each run ends only when the global optimal cost is achieved. The instance att532 only have two global optima, and the distance between those two global optima is 2, which means there are totally 534 different edges in the two global optima. Fig. \ref{fig:att532GOE} shows the 534 edges that belong to the two global optima of att532. The instance u2319 has more than \text{2,000} global optima and the mean distance among them is large, hence the edge number in the global optima set of u2319 is huge and hard to be visualized.

\subsection{Experiment Settings}
In our experiment, we execute \text{1,000} runs of GLS and \text{1,000} runs of EB-GLS. Each GLS run has a corresponding EB-GLS run which starts from the same initial solution. All the runs of GLS and EB-GLS stop only when a globally optimum is found. In both EB-GLS and GLS, the FLS strategy and the first-improvement LS strategy are applied. In EB-GLS implementation, the GLS is run for the first \text{10,000} iterations and then EB-GLS starts. $s_e$ is updated once every 100 iterations. We have recorded the best solution found so far with regards to $g$ after every \text{1,000} iterations in each run of these two algorithms.

\subsection{Performance Metrics}
On the instance att532, we measure the search performance of both GLS and EB-GLS by the following four metrics:
\begin{itemize}
  \item \textbf{Average Best Excess $\bar\epsilon$:} At the end of \text{1,000}$\times j$-th iteration in run $k$, the best excess $\epsilon_{j,k}$ is the excess of the best solution found so far:
      \begin{equation}\label{eq:excess}
        \begin{array}{l}
          \epsilon_{j,k} = \\
          \frac{\mbox{current best solution cost} - \mbox{globally optimal cost}}{\mbox{globally optimal cost}}\\
          \times 100\%.
        \end{array}
      \end{equation}
      The average best excess among all the 1,000 runs, $\bar{\epsilon}_j$, is defined as
      \begin{equation}\label{eq:mean_excess_curve}
        \bar{\epsilon}_j = \frac{1}{1000}\sum\limits_{k=1}^{1000}\epsilon_{j,k}.
      \end{equation}
      When run $k$ has found a global optimum with less than $\text{1,000}\times j$ iterations,  $\epsilon_{j,k}$ is set to be zero.
  \item \textbf{Average Distance to Global Optimum $\bar{d}$:} At the end of \text{1,000}$\times j$-th iteration in run $k$, $d_{j,k}$ is the distance (i.e., the number of different edges) between the best solution found so far and the final global optimum found in this run. The average distance $d_j$ over \text{1,000} runs is defined as:
      \begin{equation}\label{eq:mean_d_curve}
        \bar{d}_j = \frac{1}{1000}\sum\limits_{k=1}^{1000}d_{j,k}.
      \end{equation}
      Similar to the definition of $\bar{\epsilon}_j$, $d_{j,k}$ is set to be zero if run $k$ has found a global optimum with less than $\text{1,000}\times j$ iterations.
  \item \textbf{Average Ratio of Undesirable Penalties $\bar{r}$:} At each iteration in every run, each edge has a penalty value. In total, there are 534 different edges in the two global optima of att532. It is undesirable to penalize these edges. $r_{j,k}$ is defined as the ratio between the total  penalty on those 534 edges over the total penalty on all the edges at the $\text{1,000}\times j$-th iteration in run $k$. In each algorithm,  $\bar{r}_j$ is the average of all the $r_{j,k}$ values in all the runs which have not found a global optimum at the end of $\text{1,000} \times j$-th iteration. $\bar{r}$ reflects the probability of GLS/EB-GLS penalizing the 534 edges of the global optima. Obviously, everything being equal, a lower $\bar{r}$ value reflects a more effective penalizing mechanism.
  \item \textbf{Average Ratio of Increased Undesirable Penalties $\bar{r}_{\scriptscriptstyle\Delta}$:} From the $\text{1,000}\times (j-1)$-th iteration to the $\text{1,000}\times j$-th iteration in run $k$, $r_{\scriptscriptstyle\Delta,j,k}$ is defined as the ratio between the increment of the total penalty on those 534 edges over the increment of the total penalty on all the edges during that \text{1,000} iterations. $\bar{r}_{\scriptscriptstyle\Delta,j}$ is the average of all the $r_{\scriptscriptstyle\Delta,j,k}$ values in all the runs which have not found a global optimum at the end of $\text{1,000} \times j$-th iteration. $\bar{r}_{\scriptscriptstyle\Delta}$ reflects the probability of GLS/EB-GLS penalizing the 534 edges of the global optima during every \text{1,000} iterations.
\end{itemize}

Since the number of global optima in u2319 is huge, and the number of the edges in the global optima is also huge, we only calculate the average best excess $\bar\epsilon$ in the experiment on u2319.

\subsection{Experiment Results}
\subsubsection{Results on the instance with big valley structure: att532}
\begin{figure}
  \centering
    \includegraphics[width=\textwidth]{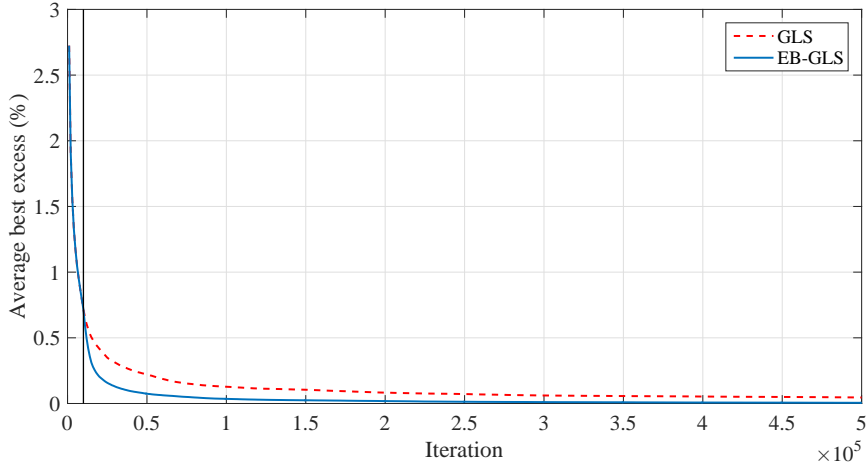}\\
  \caption{The average best excess $\bar \epsilon$ of GLS and EB-GLS on the instance att532. $\bar \epsilon$ (vertical axis) is plotted against the iteration (horizontal axis)}\label{fig:att532_excess} 
\end{figure}

\begin{figure}
  \centering
    \includegraphics[width=\textwidth]{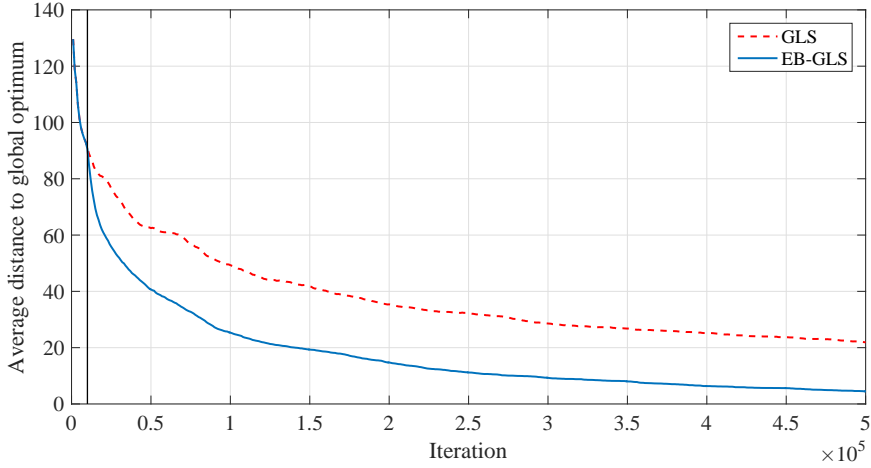}\\
  \caption{The average distance to global optimum $\bar{d}$ of GLS and EB-GLS on the instance att532. $\bar{d}$ (vertical axis) is plotted against the iteration (horizontal axis)}\label{fig:att532_dist} 
\end{figure}

Fig. \ref{fig:att532_excess} shows on att532 how the average best excess $\bar \epsilon$ of GLS and EB-GLS changes with the iteration. The \text{10,000}th iteration, which is marked by a vertical solid line in Fig. \ref{fig:att532_excess}, is the time when EB-GLS starts its own penalizing mechanism. Fig. \ref{fig:att532_dist} shows how the average distance to global optimum $\bar{d}$ changes with the iteration in GLS and EB-GLS respectively.

From Fig. \ref{fig:att532_excess} and Fig. \ref{fig:att532_dist}, it is clear that after EB-GLS starts using its penalty mechanism at the \text{10,000}th iteration,  EB-GLS performs much better than GLS in terms of both $\bar \epsilon$ and $\bar d$. It implies that the best solutions found by the EB-GLS runs have lower costs and are closer to the global optimum.  EB-GLS can always keep its superiority.

\begin{figure}
  \centering
  \subfigure[$\bar r_{\scriptscriptstyle{\mathrm{EB\text{-}GLS}}}$ and $\bar r_{\scriptscriptstyle{\mathrm{GLS}}}$]{
  \label{fig:pen_ratio_50}
  \includegraphics[width=\textwidth]{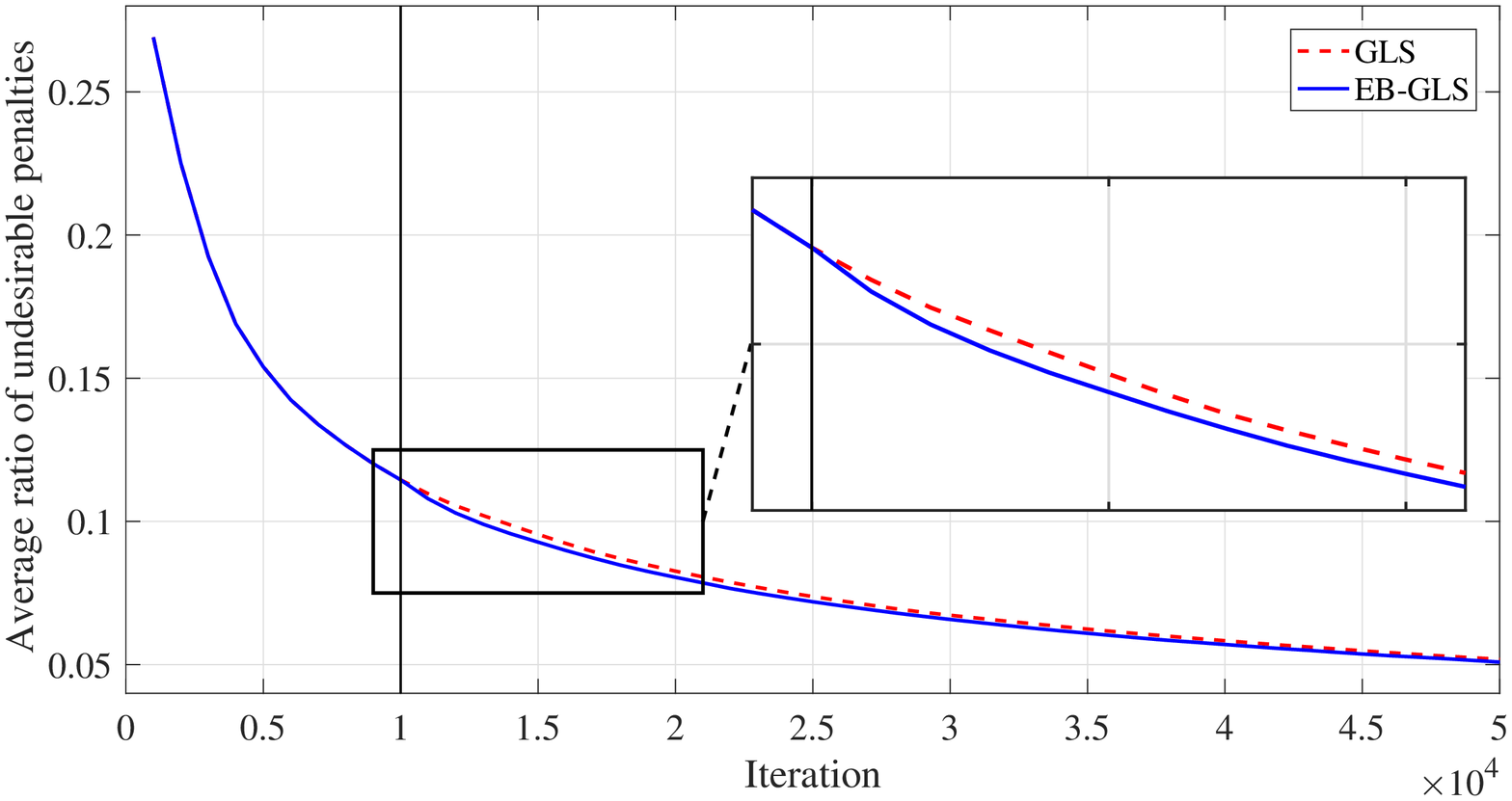}}
  \subfigure[$\bar r_{\scriptscriptstyle{\mathrm{EB\text{-}GLS}}} - \bar r_{\scriptscriptstyle{\mathrm{GLS}}}$]{
  \label{fig:pen_ratio_diff_50}
  \includegraphics[width=\textwidth]{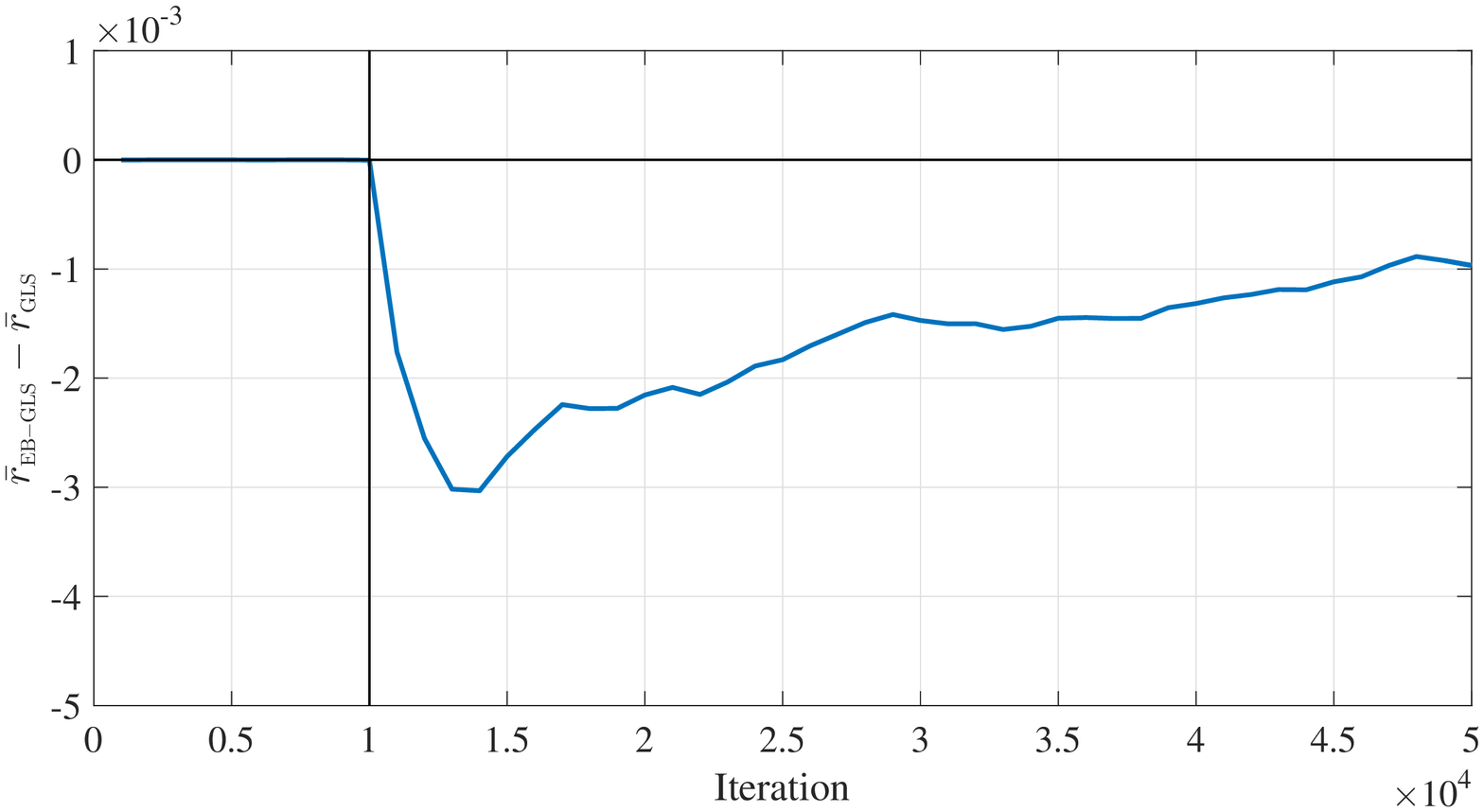}}
  \caption{The average ratio of undesirable penalties $\bar{r}$ of GLS and EB-GLS on the instance att532. (a): $\bar r$ is plotted against the iteration (horizontal axis). (b): $\bar r_{\scriptscriptstyle{\mathrm{EB\text{-}GLS}}} - \bar r_{\scriptscriptstyle{\mathrm{GLS}}}$ (vertical axis) is plotted against the iteration (horizontal axis)}\label{fig:pen_ratio_50_all} 
\end{figure}

Fig. \ref{fig:pen_ratio_50_all} shows how the average ratio of undesirable penalties $\bar r$ changes with the iteration in GLS and EB-GLS respectively. It shows the change over the first \text{50,000} iterations. All of the \text{1,000} GLS runs and 935 EB-GLS runs (the other runs have terminated within 50,000 iterations) are used to calculate $\bar r$ in the first \text{50,000} iterations. It is obvious that $\bar r$ decreases in both EB-GLS and GLS, which suggests that both algorithms are able to identify good features.
However, Fig. \ref{fig:pen_ratio_50} does not show any significant difference between EB-GLS and GLS in terms of $\bar r$. It is because $\bar r$ is calculated based on the accumulated penalties. To see the difference more clearly, Fig. \ref{fig:pen_ratio_diff_50} plots the difference between these two values. We can observe from it that after the \text{10,000}th iteration, the $\bar r$ value of EB-GLS is smaller than the $\bar r$ value of GLS. This means that EB-GLS is less likely to penalize the edges of the global optima compared to GLS.

\begin{figure}
  \centering
  \subfigure[$\bar r_{\scriptscriptstyle{\Delta,\mathrm{EB\text{-}GLS}}}$ and $\bar r_{\scriptscriptstyle{\Delta,\mathrm{GLS}}}$]{
  \label{fig:inc_pen_ratio_50}
  \includegraphics[width=\textwidth]{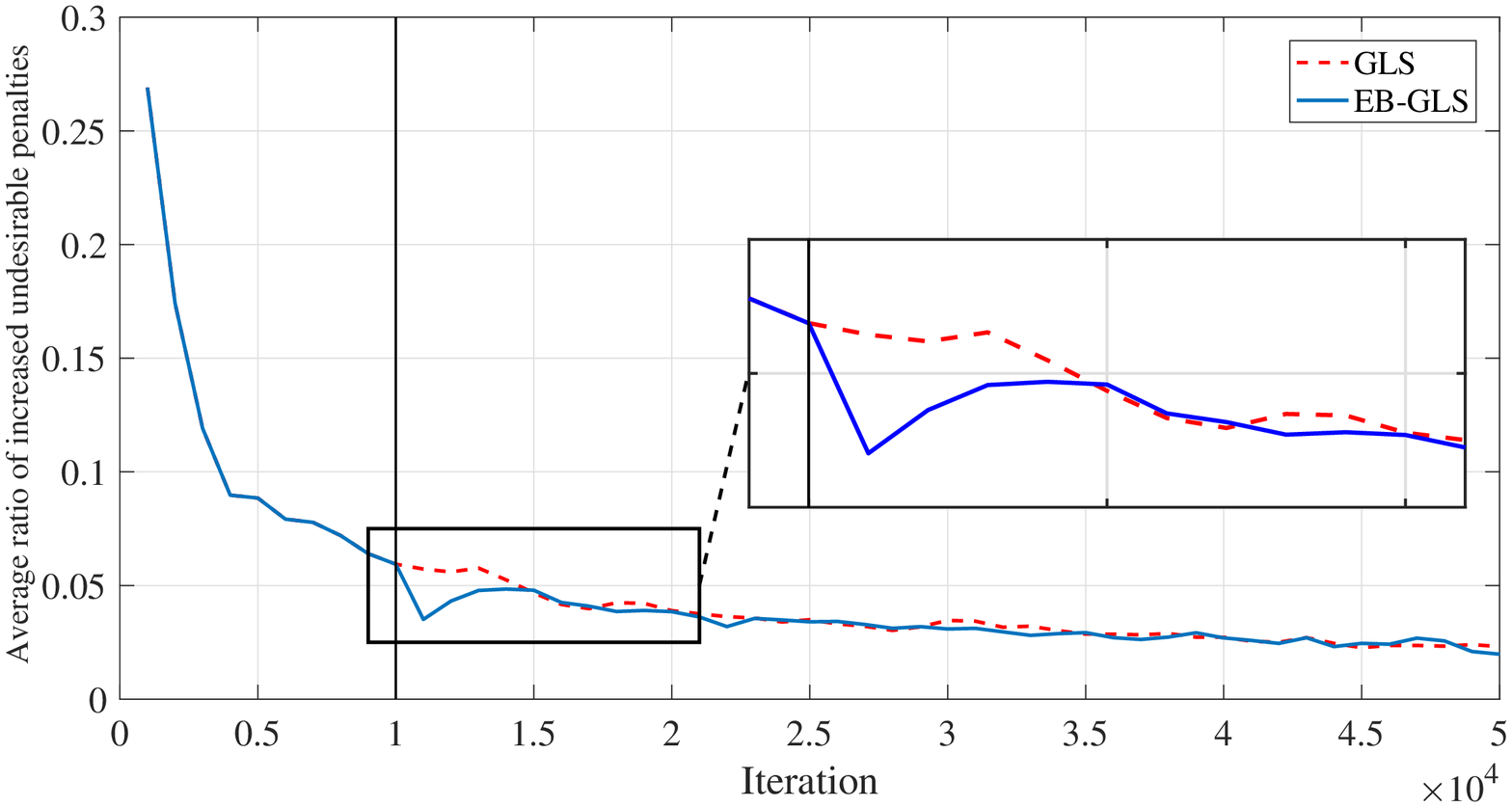}}
  \subfigure[$\bar r_{\scriptscriptstyle{\Delta,\mathrm{EB\text{-}GLS}}} - \bar r_{\scriptscriptstyle{\Delta,\mathrm{GLS}}}$]{
  \label{fig:inc_pen_ratio_diff_50}
  \includegraphics[width=\textwidth]{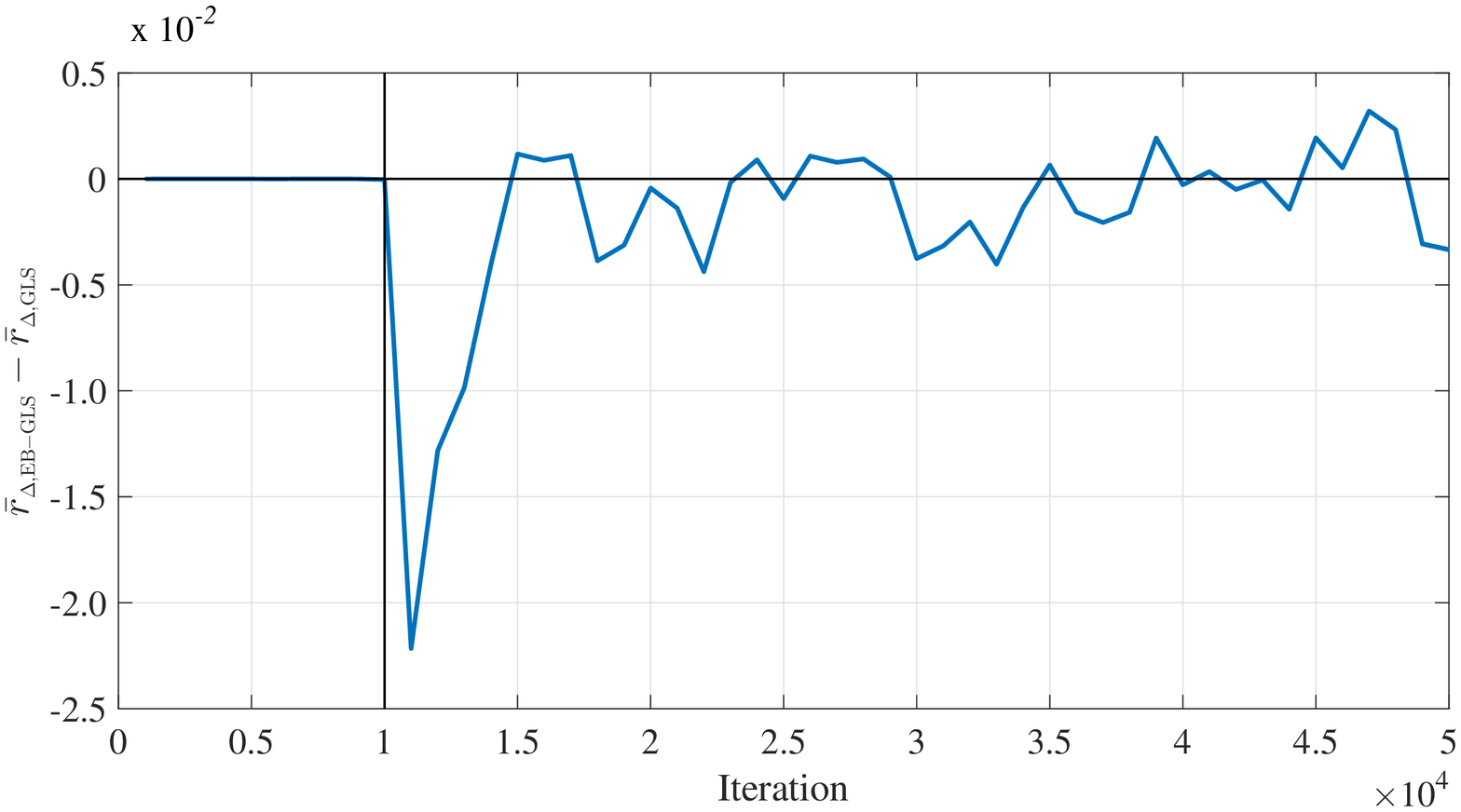}}
  \caption{The average ratio of increased undesirable penalties $\bar{r}_{\scriptscriptstyle\Delta}$ of GLS and EB-GLS on the instance att532. (a): $\bar r_{\scriptscriptstyle\Delta}$ is plotted against the iteration (horizontal axis). (b): $\bar r_{\scriptscriptstyle{\Delta,\mathrm{EB\text{-}GLS}}} - \bar r_{\scriptscriptstyle{\Delta,\mathrm{GLS}}}$ (vertical axis) is plotted against the iteration (horizontal axis)}\label{fig:inc_pen_ratio_50_all} 
\end{figure}

Fig. \ref{fig:inc_pen_ratio_50_all} shows how $\bar r_{\scriptscriptstyle\Delta}$ changes with the iteration in GLS and EB-GLS separately.
From Fig. \ref{fig:inc_pen_ratio_50} and \ref{fig:inc_pen_ratio_diff_50}, it is evident that, after the \text{10,000}th iteration, EB-GLS penalizes less global optima's edges than GLS. Especially at the \text{11,000}th iteration, the $\bar r_{\scriptscriptstyle\Delta}$ value of EB-GLS is 38\% smaller than  that of GLS. However, after the \text{11,000}th iteration, the $\bar r_{\scriptscriptstyle\Delta}$ value of EB-GLS increases until it reaches the same level of GLS's $\bar r_{\scriptscriptstyle\Delta}$ value. It is because, as the search progresses, the solutions found by EB-GLS is closer to the global optima than that of GLS (as shown in Fig. \ref{fig:att532_dist}). This means the current solution of EB-GLS contains more common edges with the global optima than that of GLS. Since GLS/EB-GLS only penalizes the edges in its current solution, the chances of EB-GLS penalizing the edges in the global optima increases. It makes EB-GLS's $\bar r_{\scriptscriptstyle\Delta}$ increase. After the \text{15,000}th iteration, the increase trend ends and EB-GLS's $\bar r_{\scriptscriptstyle\Delta}$ is about the same as in GLS. However, considering that the current solution of EB-GLS contains more edges of the global optima than that of GLS, the similarity of the $\bar r_{\scriptscriptstyle\Delta}$ values of EB-GLS and GLS confirms the advantage of EB-GLS over GLS.

\begin{figure}
  \centering
  \includegraphics[width=0.8\textwidth]{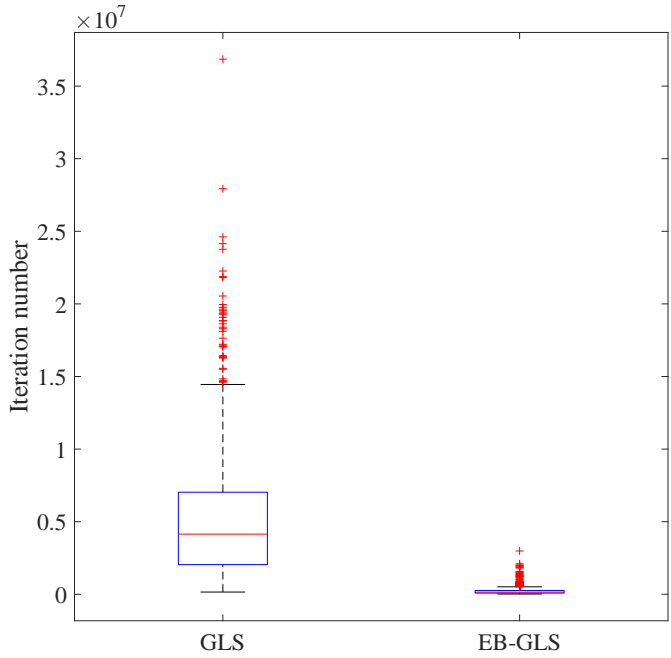}\\
  \caption{The box plot of the iteration number GLS and EB-GLS take to find a global optimum on the instance att532}\label{fig:boxplot_iter_att532} 
\end{figure}

In summary, the design goal of EB-GLS, which is to exploit the big valley structure in order to reduce the probability of penalizing the edges of global optima, has been achieved. Although the probability of EB-GLS penalizing the edges in the global optima is only a little smaller than that of GLS (as shown in Fig. \ref{fig:pen_ratio_50}), this little difference endows EB-GLS with a significantly better performance. Fig. \ref{fig:boxplot_iter_att532} is the box plot of the number of iterations GLS and EB-GLS take to find a global optimum on the instance att532. Based on the Mann-Whitnet U-test with a 0.05 significance level, we conclude that the total iteration number of EB-GLS is significantly smaller than that of GLS, which means that EB-GLS is significantly faster than GLS.

\subsubsection{Results on the instance without big valley structure: u2319}
\begin{figure}
  \centering
  \includegraphics[width=\textwidth]{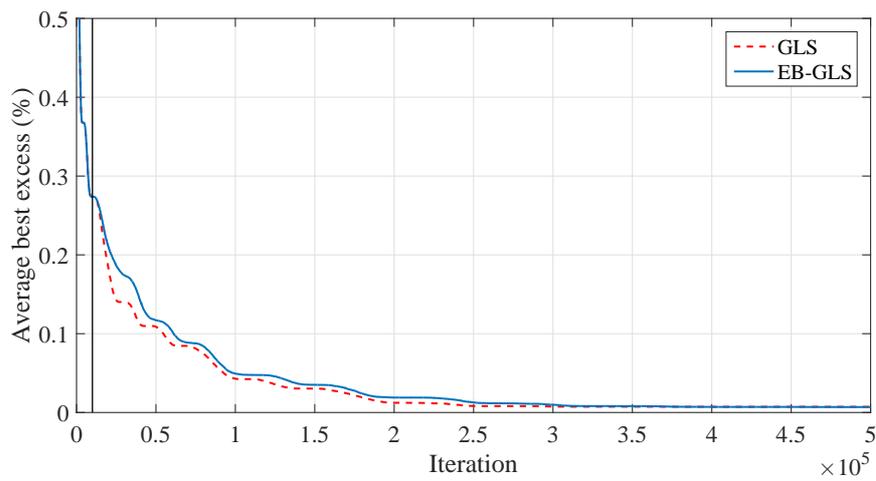}\\
  \caption{The average best excess, $\bar \epsilon$, in GLS and EB-GLS. $\bar \epsilon$ (vertical axis) is plotted against the iteration (horizontal axis). Instance: u2319}\label{fig:u2319_excess_500}  
\end{figure}

\begin{figure}
  \centering
  \includegraphics[width=0.8\textwidth]{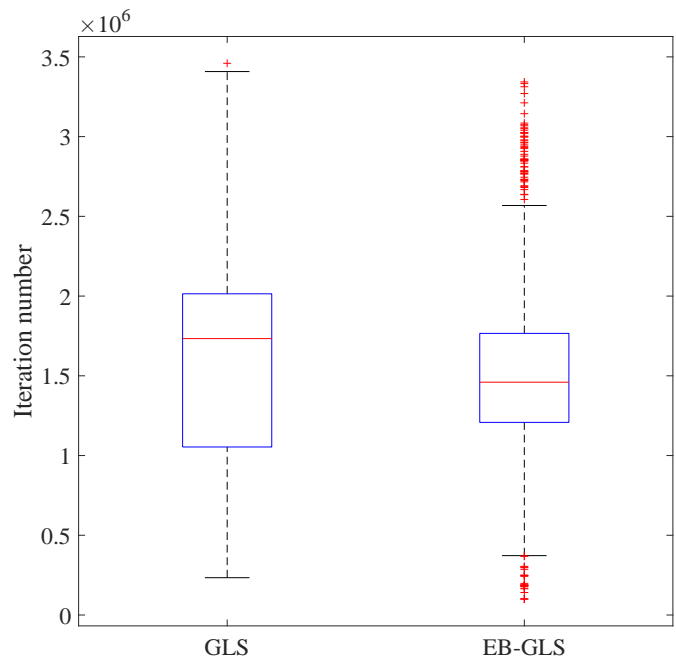}\\
  \caption{The box plot of the iteration number GLS and EB-GLS take to find a global optimum on the instance u2319}\label{fig:boxplot_iter_u2319} 
\end{figure}

Fig. \ref{fig:u2319_excess_500} shows how the average best excess $\bar \epsilon$ changes with the iteration on u2319. We can see that, GLS's $\bar \epsilon$ value is lower than that of EB-GLS in the early phase of the search process. But as the search continues, the gap between the two curves become small. Fig. \ref{fig:boxplot_iter_u2319} is the box plot of the number of iterations GLS and EB-GLS take to find a global optimum on the instance u2319. In Fig. \ref{fig:boxplot_iter_u2319}, the average iteration number GLS takes to find a global optimum is \text{1,643,358}, while the average iteration number EB-GLS takes to find a global optimum is \text{1,481,367}. We can see that, on the instance u2319 that does not have a big valley structure, the performance of EB-GLS does not drastically deteriorate. 

Having say that, this experiment only proves that EB-GLS out-performs GLS on att532, and EB-GLS's performance does not drastically deteriorate on u2319. In the next section, we will compare the performances of EB-GLS and GLS on more instances, so as to fully assess the performance of EB-GLS against GLS.

\section{Performance Comparison} \label{sec:perform_compare}
In this paper, we argue that the original GLS can be improved by exploiting the big valley structure. To prove the effectiveness of our idea, in the following part, we compare the performance of EB-GLS and GLS on a large number of symmetric TSP instances. Our objective is not to obtain a best algorithm for the TSP, hence here we do not compare EB-GLS with the state-of-the-art TSP algorithms, such as LKH~\cite{helsgaun2000effective}. Comparative experiments are conducted on two kinds of instances: instances from TSPLIB and randomly generated instances. Both EB-GLS and GLS are implemented in GNU C++ with O2 optimizing compilation. The computing platform is two 6-core 2.00GHz Intel Xeon E5-2620 CPUs (24 Logical Processors) under CentOS 6.4.

\subsection{Comparison on TSPLIB Instances} \label{sec:cmp_109}
There are 111 symmetric TSP instances in TSPLIB. In this experiment we choose 109 instances as test instances. The instance linhp312 is excluded because it requests the solution to contain a fixed edge. The largest instance pla85900 is excluded because of the RAM limitation of our hardware.

\subsubsection{Experiment settings}
Most of the settings of this comparative experiment remain the same as the settings of the experiment introduced in Section \ref{sec:EBGLS_behavior}, except the following:

\begin{itemize}
  \item We execute GLS and EB-GLS 100 runs on each instance.
  \item Both algorithms stop when a global optimum  is found or the predetermined maximum runtime is reached. For a TSP instance with $N$ cities, the maximum runtime $T=\lceil N/10 \rceil$ seconds. For example, the maximum runtime on the instance fl1577 (which has 1577 cities) is 158 seconds.
  \item For the EB-GLS runs on the instances with not less than 1000 cities, we first run GLS for $\lfloor T/10\rfloor$ seconds and then change to EB-GLS. For example, on the instance fl1577, the starting time of applying EB-GLS in the EB-GLS runs is the $\lfloor 158/10 \rfloor = 16$th second. For the EB-GLS runs on the instances with less than 1000 cities, the EB-GLS mechanism starts at the beginning.
\end{itemize}

\subsubsection{Experiment results} \label{sec:cmp_1000}
We use three metrics to measure the performance of GLS/EB-GLS. The first metric is the number of runs that successfully find a global optimum. The second one is the average best excess of all the 100 runs. The third one is the average real runtime of all the 100 runs. Table \ref{tbl:cmprsn_rslt} shows the comparison results between EB-GLS and GLS on the 33 TSPLIB instances with not less than 1000 cities. The comparison results on the rest 76 TSPLIB instances can be found in the Appendix. In Table \ref{tbl:cmprsn_rslt} the bold font means that one algorithm gets a better metric value than the other one. We perform Mann-Whitney U-test on the excess data and the runtime data respectively. In Table \ref{tbl:cmprsn_rslt}, the sign ``$+$''(resp. ``$-$'',``$\approx$'') indicates that EB-GLS achieves better (resp. worse, equivalent) results than GLS using a Mann-Whitney U-test at the 0.05 significance level.
\begin{table}
\caption{Comparison results between EB-GLS and GLS on the TSPLIB instances with not less than 1000 cities, the better metric values are marked by bold texts}
\centering
\label{tbl:cmprsn_rslt} 
\resizebox{\textwidth}{!}{
\begin{tabular}{l|c|c|c|r|r|c|r|r|c}
\hline
\multirow{2}{*}{Instance} & \multirow{2}{*}{\begin{minipage}{28pt}\tiny{Max Runtime (s)}\end{minipage}} & \multicolumn{2}{c|}{Success of 100} & \multicolumn{2}{c|}{Average Excess (\%)} & \multirow{2}{*}{\begin{minipage}{30pt}Excess P-value\end{minipage}} & \multicolumn{2}{c|}{Average Runtime (s)} & \multirow{2}{*}{\begin{minipage}{30pt}Runtime P-value\end{minipage}} \\ 
\cline{3-4}\cline{5-6}\cline{8-9}
&& \scriptsize{GLS} & \scriptsize{EB-GLS} & \multicolumn{1}{c|}{\scriptsize{GLS}} & \multicolumn{1}{c|}{\scriptsize{EB-GLS}} && \multicolumn{1}{c|}{\scriptsize{GLS}} & \multicolumn{1}{c|}{\scriptsize{EB-GLS}} & \\
\hline
dsj1000 & 100 & 0 & \textbf{1} & 0.2821 & \textbf{0.0856} $+$ & 5.11e-34 & 100.00 & \textbf{99.99} $\approx$ & 3.22e-01  \\
pr1002 & 101 & 0 & \textbf{82} & 0.0610 & \textbf{0.0009} $+$ & 1.43e-36 & 101.00 & \textbf{60.13} $+$ & 2.69e-29  \\
si1032 & 104 & 0 & \textbf{20} & 0.0340 & \textbf{0.0139} $+$ & 1.04e-13 & 104.00 & \textbf{94.75} $+$ & 2.72e-06  \\
u1060 & 106 & 0 & \textbf{48} & 0.0720 & \textbf{0.0080} $+$ & 2.05e-34 & 106.00 & \textbf{87.70} $+$ & 4.98e-15  \\
vm1084 & 109 & 0 & \textbf{32} & 0.0638 & \textbf{0.0159} $+$ & 2.89e-31 & 109.00 & \textbf{87.72} $+$ & 9.14e-10  \\
pcb1173 & 118 & 0 & \textbf{22} & 0.0737 & \textbf{0.0095} $+$ & 5.37e-34 & 118.00 & \textbf{103.85} $+$ & 7.58e-07  \\
d1291 & 130 & 1 & \textbf{16} & 0.1196 & \textbf{0.0739} $+$ & 6.76e-07 & 128.86 & \textbf{119.78} $+$ & 1.80e-04  \\
rl1304 & 131 & 0 & \textbf{76} & 0.0443 & \textbf{0.0261} $+$ & 2.19e-15 & 131.00 & \textbf{69.14} $+$ & 2.00e-26  \\
rl1323 & 133 & 1 & \textbf{26} & 0.0748 & \textbf{0.0387} $+$ & 3.91e-09 & 132.82 & \textbf{116.38} $+$ & 2.16e-07  \\
nrw1379 & 138 & 0 & \textbf{7} & 0.0896 & \textbf{0.0164} $+$ & 4.33e-34 & 138.00 & \textbf{134.60} $+$ & 7.31e-03  \\
fl1400 & 140 & 0 & \textbf{1} & 0.5054 & \textbf{0.3119} $+$ & 1.44e-19 & 140.00 & \textbf{139.79} $\approx$ & 3.22e-01  \\
u1432 & 144 & 53 & \textbf{93} & 0.0092 & \textbf{0.0020} $+$ & 9.41e-10 & 112.79 & \textbf{67.09} $+$ & 1.69e-12  \\
fl1577 & 158 & 0 & 0 & 0.3325 & \textbf{0.2555} $+$ & 2.50e-07 & 158.00 & 158.00 $\approx$ & -  \\
d1655 & 166 & 0 & \textbf{5} & 1.0514 & \textbf{0.9681} $+$ & 2.46e-03 & 166.00 & \textbf{163.25} $+$ & 2.42e-02  \\
vm1748 & 175 & 0 & \textbf{9} & 0.1350 & \textbf{0.0496} $+$ & 2.17e-31 & 175.00 & \textbf{171.31} $+$ & 2.23e-03  \\
u1817 & 182 & 0 & \textbf{1} & 0.2220 & \textbf{0.1398} $+$ & 1.01e-16 & 182.00 & \textbf{181.93} $\approx$ & 3.22e-01  \\
rl1889 & 189 & 0 & \textbf{3} & 0.2011 & \textbf{0.0753} $+$ & 1.24e-24 & 189.00 & \textbf{187.30} $\approx$ & 8.27e-02  \\
d2103 & 211 & 0 & 0 & 0.1728 & \textbf{0.1339} $+$ & 6.09e-09 & 211.00 & 211.00 $\approx$ & -  \\
u2152 & 216 & 0 & 0 & 0.2776 & \textbf{0.1725} $+$ & 1.88e-15 & 216.00 & 216.00 $\approx$ & -  \\
u2319 & 232 & \textbf{30} & 6 & \textbf{0.0051} & 0.0068 $-$ & 1.06e-05 & \textbf{209.62} & 227.16 $-$ & 1.43e-05  \\
pr2392 & 240 & 0 & 0 & 0.1429 & \textbf{0.0410} $+$ & 1.73e-26 & 240.00 & 240.00 $\approx$ & -  \\
pcb3038 & 304 & 0 & 0 & 0.2025 & \textbf{0.0732} $+$ & 8.08e-34 & 304.00 & 304.00 $\approx$ & -  \\
fl3795 & 380 & 0 & 0 & \textbf{2.1908} & 2.4098 $\approx$ & 8.87e-01 & 380.00 & 380.00 $\approx$ & -  \\
fnl4461 & 447 & 0 & 0 & 0.2547 & \textbf{0.1175} $+$ & 2.79e-34 & 447.00 & 447.00 $\approx$ & -  \\
rl5915 & 592 & 0 & 0 & 0.5100 & \textbf{0.3941} $+$ & 1.92e-06 & 592.00 & 592.00 $\approx$ & -  \\
rl5934 & 594 & 0 & 0 & 0.7126 & \textbf{0.6214} $+$ & 3.53e-03 & 594.00 & 594.00 $\approx$ & -  \\
pla7397 & 740 & 0 & 0 & 0.4588 & \textbf{0.3834} $+$ & 1.27e-07 & 740.00 & 740.00 $\approx$ & -  \\
rl11849 & 1185 & 0 & 0 & 0.8366 & \textbf{0.7266} $+$ & 2.24e-06 & 1185.00 & 1185.00 $\approx$ & -  \\
usa13509 & 1351 & 0 & 0 & 0.8776 & \textbf{0.6231} $+$ & 7.17e-26 & 1351.00 & 1351.00 $\approx$ & -  \\
brd14051 & 1406 & 0 & 0 & 1.8110 & \textbf{1.6733} $+$ & 2.51e-06 & 1406.00 & 1406.00 $\approx$ & -  \\
d15112 & 1512 & 0 & 0 & 0.7742 & \textbf{0.5720} $+$ & 2.68e-27 & 1512.00 & 1512.00 $\approx$ & -  \\
d18512 & 1852 & 0 & 0 & 0.8098 & \textbf{0.7522} $+$ & 2.13e-09 & 1852.00 & 1852.00 $\approx$ & -  \\
pla33810 & 3381 & 0 & 0 & \textbf{1.2438} & 1.3276 $-$ & 3.18e-02 & 3381.00 & 3381.00 $\approx$ & -  \\
\hline
\end{tabular}
}
\end{table}

To compare the performances between GLS and EB-GLS, we use the ``domination'' concept from multi-objective optimization to judge which one is better. Through the statistical tests, if algorithm A is not significantly worse than algorithm B on all three metrics, and A is significantly better than B on at least one metric, then we state that A out-performs B. Among the 109 instances, EB-GLS out-performs GLS on 82 instances (75.23\%), while GLS out-performs EB-GLS on only 6 instances (5.50\%). Furthermore, among the 71 instances whose city numbers are not less than 150, EB-GLS out-performs GLS on 65 instances (91.55\%), while GLS out-performs EB-GLS on only 2 instances (2.82\%), u2319 and pla33810. In Section \ref{sec:big_valley} we know that the instances u2319 does not have the big valley structure, which explains why the performance of EB-GLS is worse than that of GLS on u2319. The other instance pla33810 contains a large number of nodes, hence it is difficult to investigate its landscape structure experimentally. In summary, from the above comparison results we can conclude that the overall performance of EB-GLS is better than the performance of GLS on most instances of TSPLIB.

\subsection{Comparison on Randomly Generated Instances}
In this experiment, GLS and EB-GLS is each executed on 10 randomly generated symmetric TSP instances.
\subsubsection{Experiment settings}
\begin{figure}
  \centering
  \includegraphics[width=3.3in]{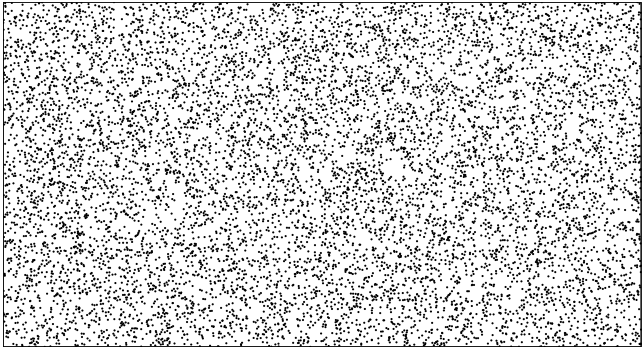}\\
  \caption{City distribution in randc10000}\label{fig:randc10000}
\end{figure}

The 10 randomly generated instances all have \text{10,000} cities. These instances are named ``randa10000'', ``randb10000'', \ldots, ``randj10000'', respectively. For each instance, both the length and width of the map are randomly uniform number generated from the interval $(1\times10^5,1.1\times10^6)$.  the positions of 10,000 cities are uniformly distributed in the map. For example, Fig. \ref{fig:randc10000} shows the city distribution in randc10000.

For each run on each test instance, the maximum runtime is \text{1,000} seconds, and in EB-GLS the starting time of applying EB-GLS is the 100th second. Since the global optima of these random instances are unknown, all runs stop only when the maximum runtime is reached. After that, the cost of the best solution found is recorded. The other experiment settings are same as that in the comparative experiment on TSPLIB instances.

\subsubsection{Experiment results}
Table \ref{tbl:cmprsn_rslt_rand} shows the average best costs got by GLS and EB-GLS, in which the smaller cost values are in bold. The sign ``+'' indicates that EB-GLS significantly out-perform GLS using a Mann-Whitney U-test at the 0.05 significance level.
\begin{table}
\caption{Comparison results on randomly generated instances, the better metric values are marked by bold texts}
\centering
\label{tbl:cmprsn_rslt_rand} 
\begin{tabular}{l|c|c|c}
\hline
\multirow{2}{*}{Instance} & \multicolumn{2}{c|}{Average Best Cost} &\multirow{2}{*}{P-value} \\
\cline{2-3}
& \multicolumn{1}{c|}{GLS} & \multicolumn{1}{c|}{EB-GLS} & \\
\hline
randa10000 & 37928164 & \textbf{37849305} $+$ & 2.11e-20  \\
randb10000 & 59934293 & \textbf{59808692} $+$ & 1.94e-21  \\
randc10000 & 50040286 & \textbf{49922413} $+$ & 1.53e-27  \\
randd10000 & 22909402 & \textbf{22857558} $+$ & 8.62e-25  \\
rande10000 & 70175627 & \textbf{70019896} $+$ & 1.20e-23  \\
randf10000 & 29017812 & \textbf{28955712} $+$ & 1.90e-26  \\
randg10000 & 11929810 & \textbf{11900595} $+$ & 4.11e-29  \\
randh10000 & 70408148 & \textbf{70248660} $+$ & 1.37e-25  \\
randi10000 & 44957797 & \textbf{44848043} $+$ & 1.25e-29  \\
randj10000 & 24271377 & \textbf{24207587} $+$ & 9.72e-30  \\
\hline
\end{tabular}
\end{table}
From Table \ref{tbl:cmprsn_rslt_rand} we can see that the performance of EB-GLS is significantly better than that of GLS on all the 10 randomly generated instances.

\subsection{Comparison with Ant Colony Optimization Algorithm}
In this section, we compare the proposed EB-GLS with a TSP algorithm that employs completely different mechanism, the Ant Colony Optimization (ACO) algorithm~\cite{dorigo2006ant}. ACO is widely used to solve the TSP since it is inspired by how ants find the shortest path to the food. Here we use the ACOTSP software package available at http://www.aco-metaheuristic.org/aco-code/. The comparison method is similar to the method we use in Section \ref{sec:cmp_109}, but here we only use 10 test instances.

We set the parameters of ACO based on \cite{oliveira2011detailed}, which are shown in Table \ref{tbl:para_ACO}. The test instances we select are eil101, d198, kroA200, rd400, d657, u724, pcb1173, u1817, d2103 and u2319. The maximum runtime of each instance is same to the experiment setting in Section \ref{sec:cmp_109}. The result data of EB-GLS also comes from the experiment in Section \ref{sec:cmp_109}.
\begin{table}[b]
\caption{Parameter settings of ACO}
\centering
\label{tbl:para_ACO}
\begin{tabular}{l|l|l}
\hline
Parameters & Description & Values\\
\hline
 $m_{\scriptscriptstyle{ACO}}$ & Number of ants & 25\\
 $\alpha$ & Influence of pheromone trails & 1 \\
 $\beta$ & Influence of heuristic information & 2 \\
 $\rho$ & Pheromone trail evaporation & 0.2\\
 LS & Local search & 3-Opt\\
 MMAS & MAX-MIN ant system & Apply\\
\hline
\end{tabular}
\end{table}

Table \ref{tbl:ACO_cmprsn_rslt} show the comparison results between EB-GLS and ACOTSP where the bold font means that one algorithm gets a better metric value than the other one. We conduct Mann-Whitney U-test with a 0.05 significance level on the excess data and the runtime data respectively. Based on the metric values and the Mann-Whitney U-test, we conclude that EB-GLS out-performs ACOTSP on nine instances, while ACOTSP out-performs EB-GLS on only one instances. Compared the results in Table \ref{tbl:ACO_cmprsn_rslt} and the results in Table \ref{tbl:cmprsn_rslt} (and Table \ref{tbl:cmprsn_rslt_c1} in the Appendix) we can see that, the original GLS performs worse than ACOTSP on several instances (e.g. rd400, u724 and pcb1173). Meanwhile the enhanced variant of GLS, EB-GLS, performs better than ACOTSP on those instances.
\begin{table}
\caption{Comparison results between EB-GLS and ACOTSP on TSPLIB instances, the better metric values are marked by bold texts}
\centering
\label{tbl:ACO_cmprsn_rslt} 
\resizebox{\textwidth}{!}{
\begin{tabular}{l|c|c|c|r|r|c|r|r|c}
\hline
\multirow{2}{*}{Instance} & \multirow{2}{*}{\begin{minipage}{28pt}\tiny{Max Runtime (s)}\end{minipage}} & \multicolumn{2}{c|}{Success of 100} & \multicolumn{2}{c|}{Average Excess (\%)} & \multirow{2}{*}{\begin{minipage}{30pt}Excess P-value\end{minipage}} & \multicolumn{2}{c|}{Average Runtime (s)} & \multirow{2}{*}{\begin{minipage}{30pt}Runtime P-value\end{minipage}} \\ 
\cline{3-4}\cline{5-6}\cline{8-9}
&& \scriptsize{ACOTSP} & \scriptsize{EB-GLS} & \multicolumn{1}{c|}{\scriptsize{ACOTSP}} & \multicolumn{1}{c|}{\scriptsize{EB-GLS}} && \multicolumn{1}{c|}{\scriptsize{ACOTSP}} & \multicolumn{1}{c|}{\scriptsize{EB-GLS}} & \\
\hline
eil101 & 11 & 100 & 100 & 0.0000 & 0.0000 $\approx$ & - & 0.05 & \textbf{0.02} $+$ & 1.73e-21  \\
d198 & 20 & 76 & \textbf{96} & 0.0015 & \textbf{0.0004} $+$ & 1.12e-04 & 10.03 & \textbf{4.79} $+$ & 2.26e-08  \\
kroA200 & 20 & 100 & 100 & 0.0000 & 0.0000 $\approx$ & - & 0.25 & \textbf{0.17} $+$ & 8.59e-15  \\
rd400 & 40 & 90 & \textbf{93} & 0.0007 & \textbf{0.0005} $\approx$ & 3.51e-01 & 14.34 & \textbf{6.86} $+$ & 9.82e-11  \\
d657 & 66 & 0 & 0 & 0.0557 & \textbf{0.0038} $+$ & 9.70e-21 & 66.00 & 66.00 $\approx$ & -  \\
u724 & 73 & 43 & \textbf{58} & 0.0261 & \textbf{0.0069} $+$ & 2.39e-05 & 56.85 & \textbf{50.43} $+$ & 2.76e-02  \\
pcb1173 & 118 & 14 & \textbf{22} & 0.0252 & \textbf{0.0095} $+$ & 2.10e-04 & 111.64 & \textbf{103.85} $+$ & 2.56e-02  \\
u1817 & 182 & 0 & \textbf{1} & 0.1563 & \textbf{0.1398} $\approx$ & 1.22e-01 & 182.00 & \textbf{181.93} $+$ & 2.38e-09  \\
d2103 & 211 & \textbf{8} & 0 & \textbf{0.0402} & 0.1339 $-$ & 6.36e-12 & \textbf{203.81} & 211.00 $-$ & 7.67e-04  \\
u2319 & 232 & 0 & \textbf{6} & 0.3068 & \textbf{0.0068} $+$ & 2.22e-38 & 232.01 & \textbf{227.16} $+$ & 5.06e-14  \\
\hline
\end{tabular}
}
\end{table}

\section{Conclusions} \label{sec:conclusion}
As an important part of memetic algorithm, LS usually finds locally optimal solutions. GLS is a simple yet powerful strategy to guide LS escape from locally optimal solutions in combinatorial optimization. Its success is due to its feature penalizing mechanism. It has been observed that many combinatorial optimization problems exhibit a big valley structure. This implies that solutions with higher quality have a better chance of being similar to other good solutions, even the global optima. To exploit this property, we have proposed a new variant of GLS, EB-GLS, with an improved feature penalizing mechanism. EB-GLS records and updates an elite solution, which is the best solution found so far during the search. Under the assumption that features present in the recorded elite solution are more likely to be part of a global optimum, EB-GLS reduces the probabilities of these features being penalized. In doing so, search efficiency can be significantly improved. We use symmetric TSP as the test problem to conduct experiments. Our objective is not to develop a new best algorithm for the TSP, but to show the effectiveness of our modification. We have conducted extensive experiments to study the behavior of EB-GLS on two TSP instance, att532 and u2319. The former instance has a big valley structure, while the latter instance does not. We also compare EB-GLS and GLS on a large number of test instances. Experimental results suggest that EB-GLS out-performs GLS on most instances, and EB-GLS's performance does not drastically deteriorate on the instance that does not have the big valley structure. Our work represents the first attempt of using big valley structure to improve GLS. In addition, we hope that the experimental methodology in this paper can inspire new methods to study penalization based algorithms. 


\begin{acknowledgements}
The work described in this paper was supported by a grant from ANR/RCC Joint Research Scheme sponsored by the Research Grants Council of the Hong Kong Special Administrative Region, China and France National Research Agency (Project No. A-CityU101/16).
\end{acknowledgements}


%
%

\begin{appendices}
\section*{Appendix}

\begin{figure}[H]
  \centering
  \subfigure[rd400]{
  \label{fig:cvd_rd400}
  \includegraphics[width=0.22\textwidth]{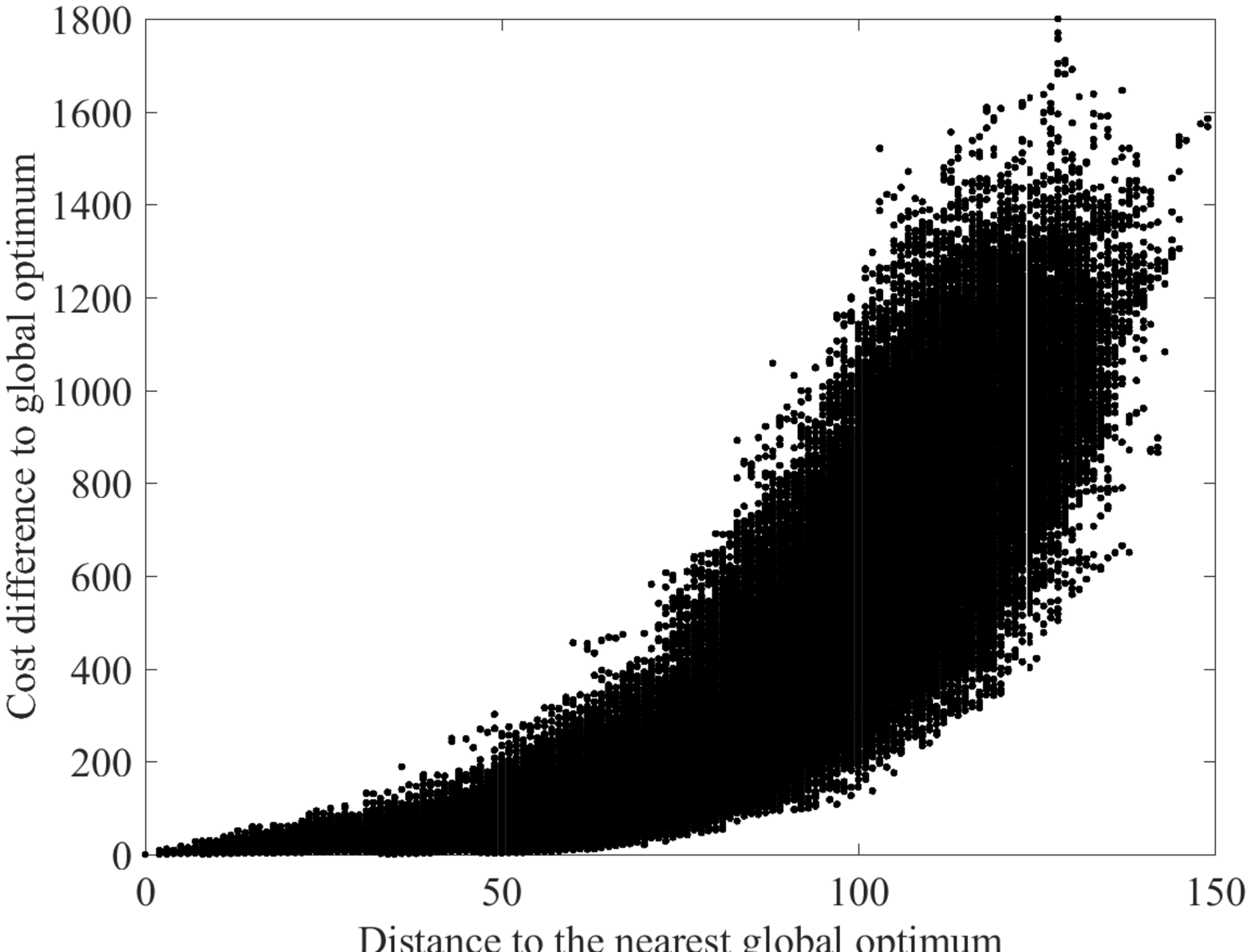}}
\hspace{0.01\textwidth}
  \subfigure[gr431]{
  \label{fig:cvd_gr431}
  \includegraphics[width=0.22\textwidth]{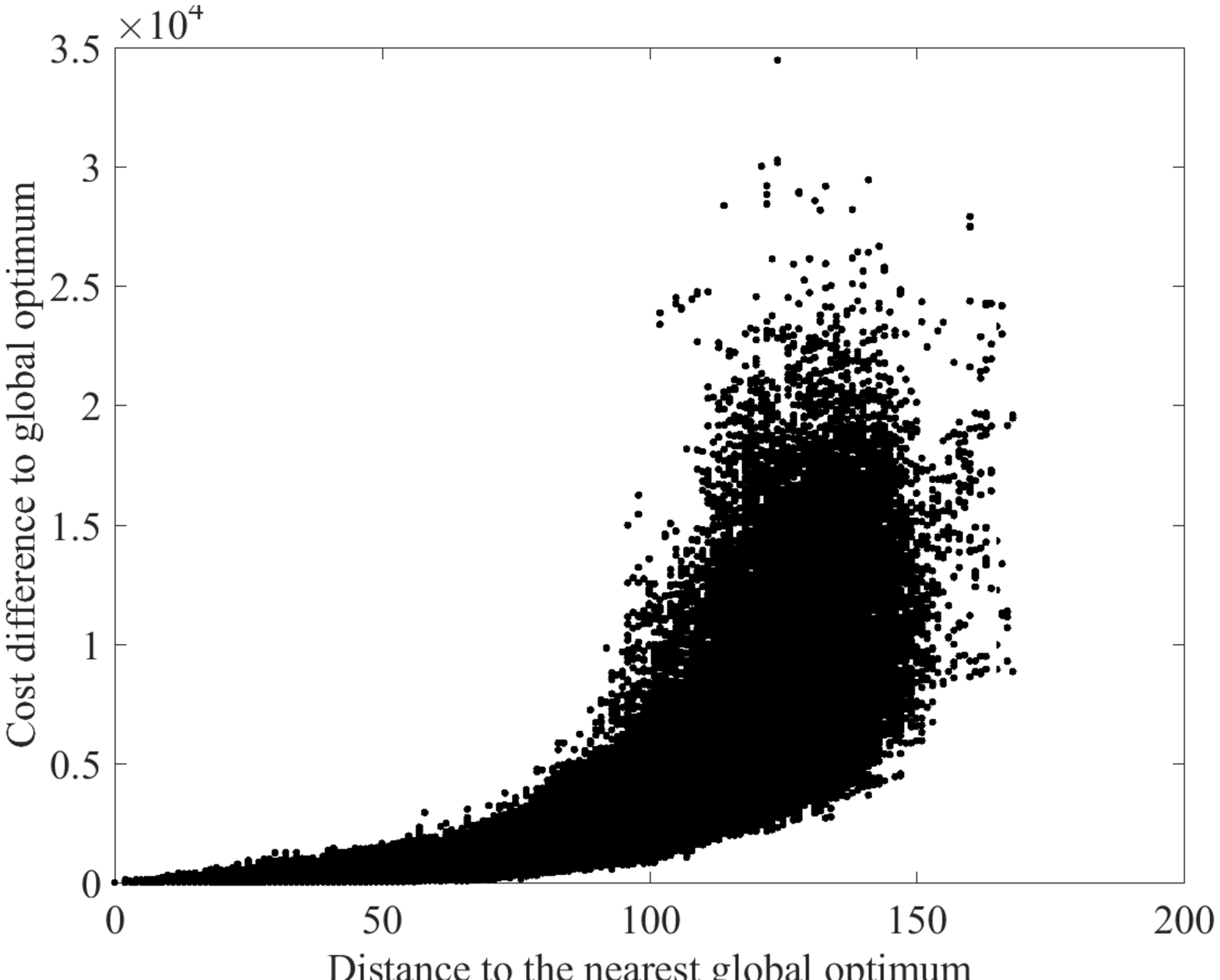}}
\hspace{0.01\textwidth}
  \subfigure[pcb442]{
  \label{fig:cvd_pcb442}
  \includegraphics[width=0.22\textwidth]{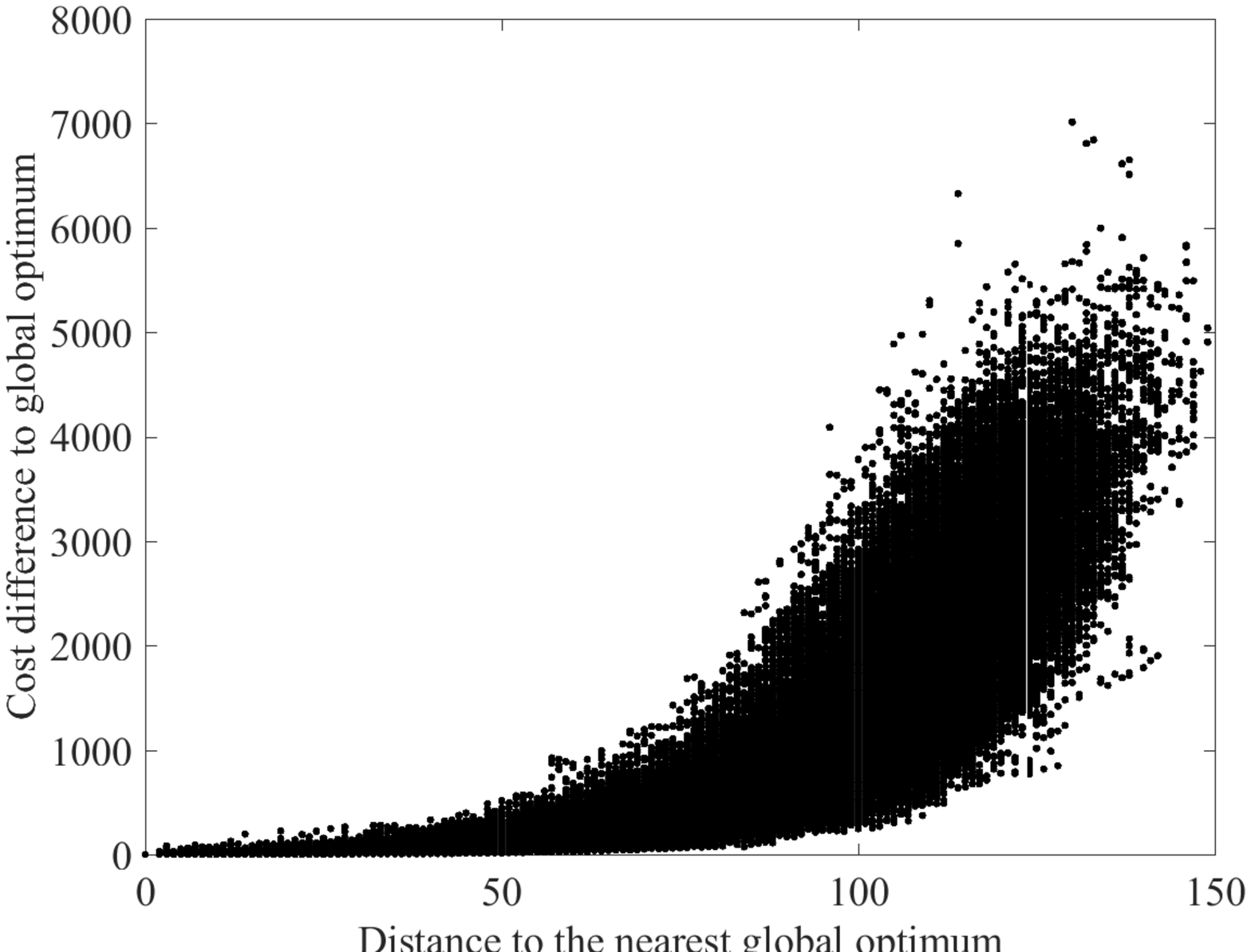}}
\hspace{0.01\textwidth}
  \subfigure[pa561]{
  \label{fig:cvd_pa561}
  \includegraphics[width=0.22\textwidth]{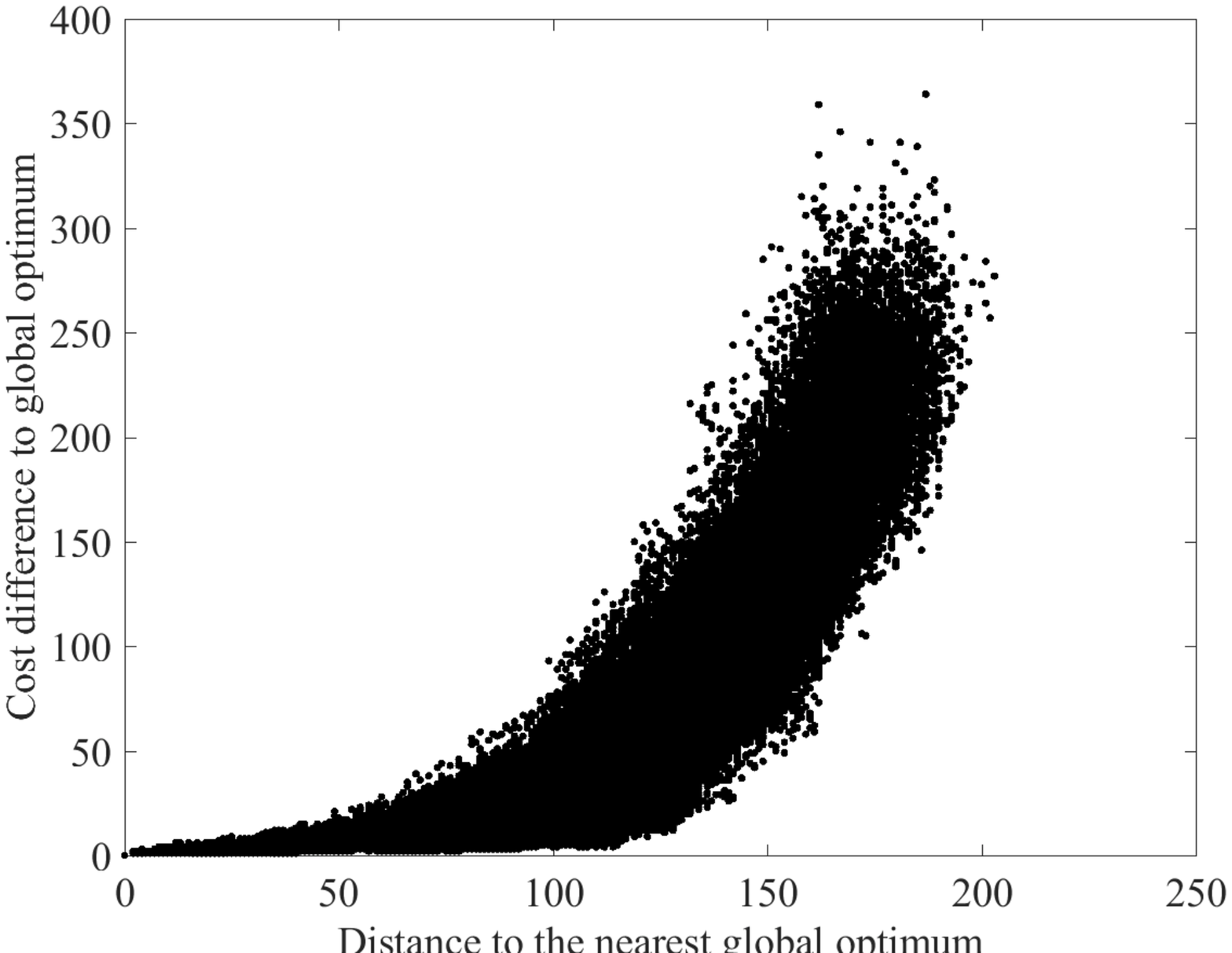}}

  \subfigure[u574]{
  \label{fig:cvd_u574}
  \includegraphics[width=0.22\textwidth]{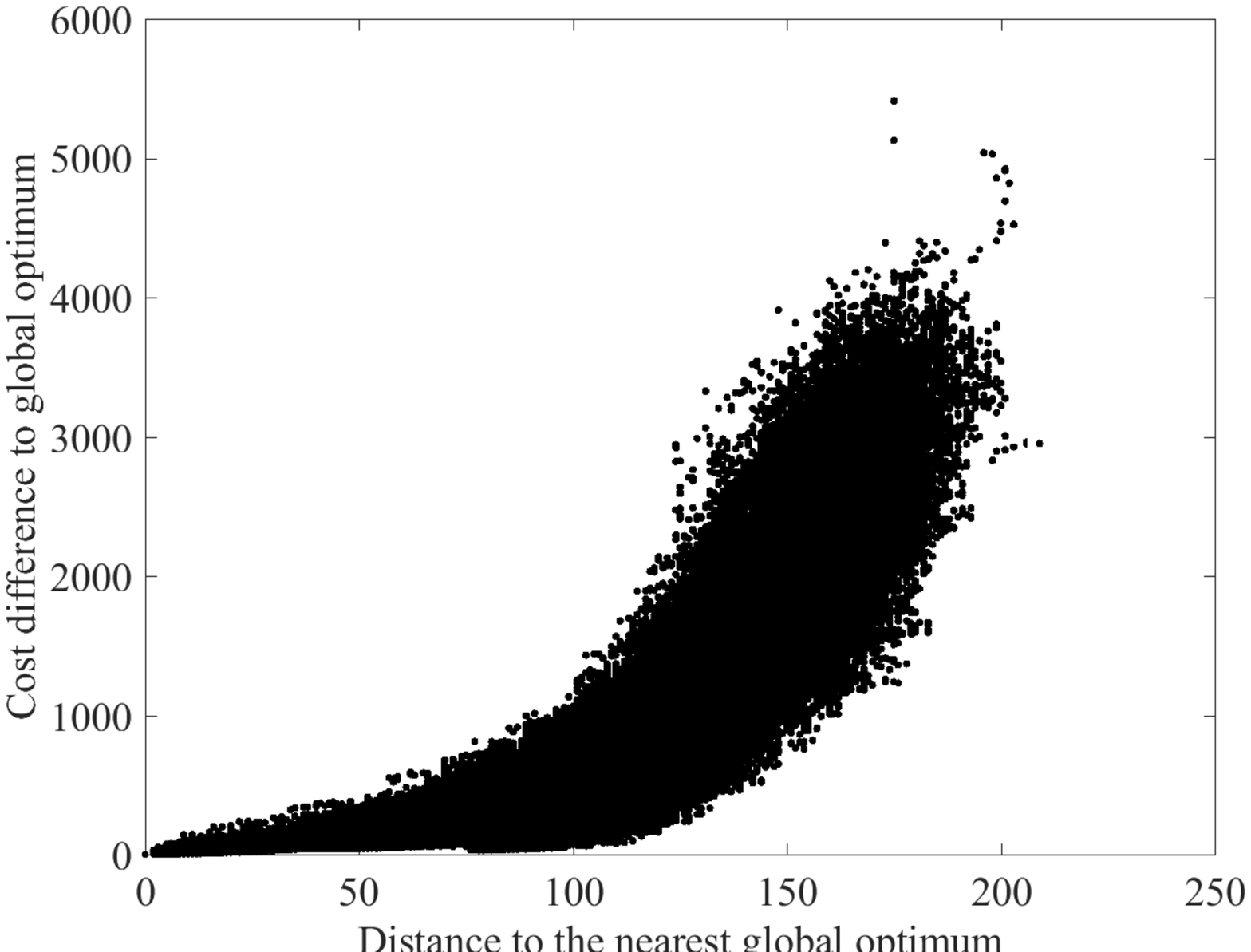}}
\hspace{0.01\textwidth}
  \subfigure[rat575]{
  \label{fig:cvd_rat575}
  \includegraphics[width=0.22\textwidth]{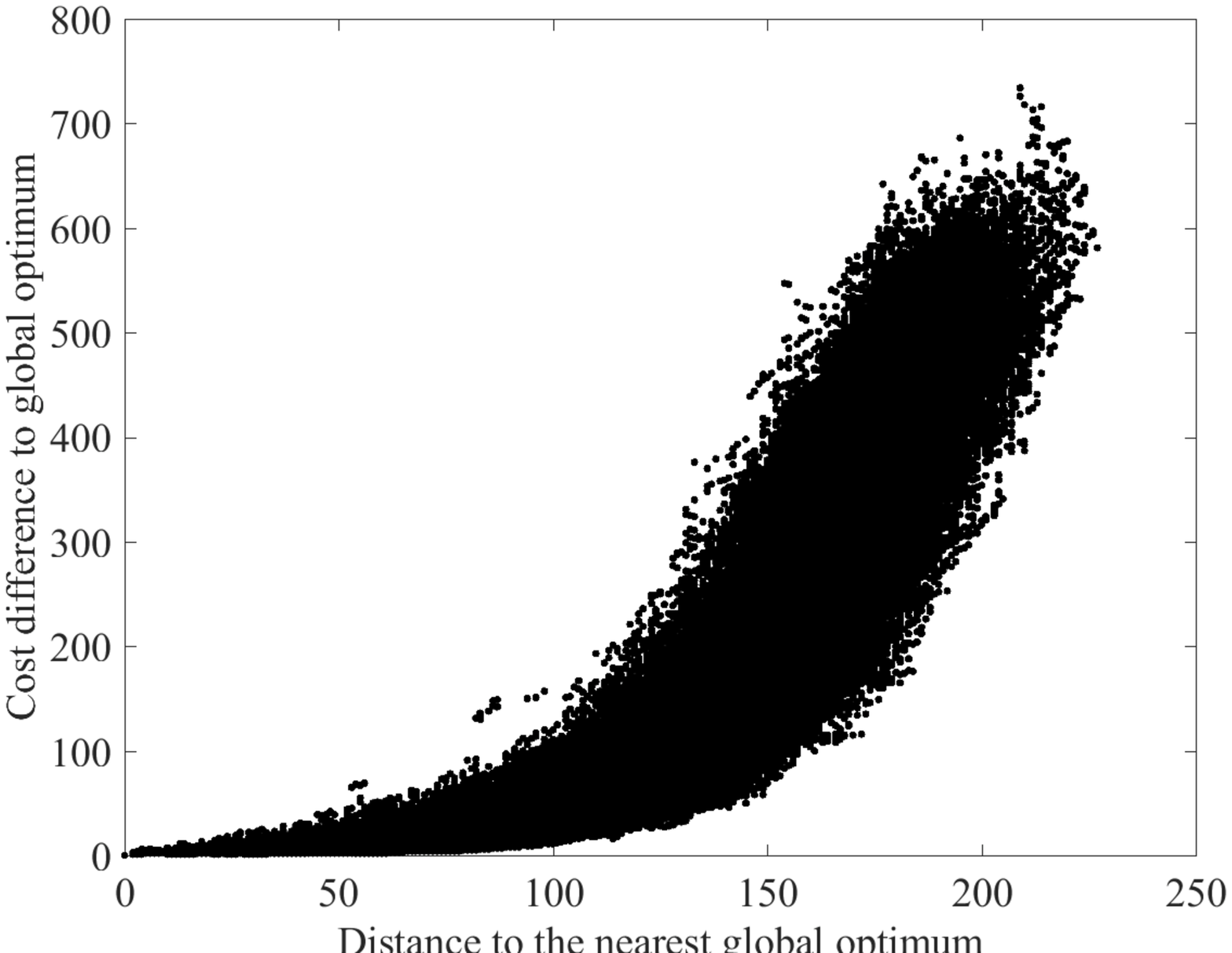}}
\hspace{0.01\textwidth}
  \subfigure[rat783]{
  \label{fig:cvd_rat783}
  \includegraphics[width=0.22\textwidth]{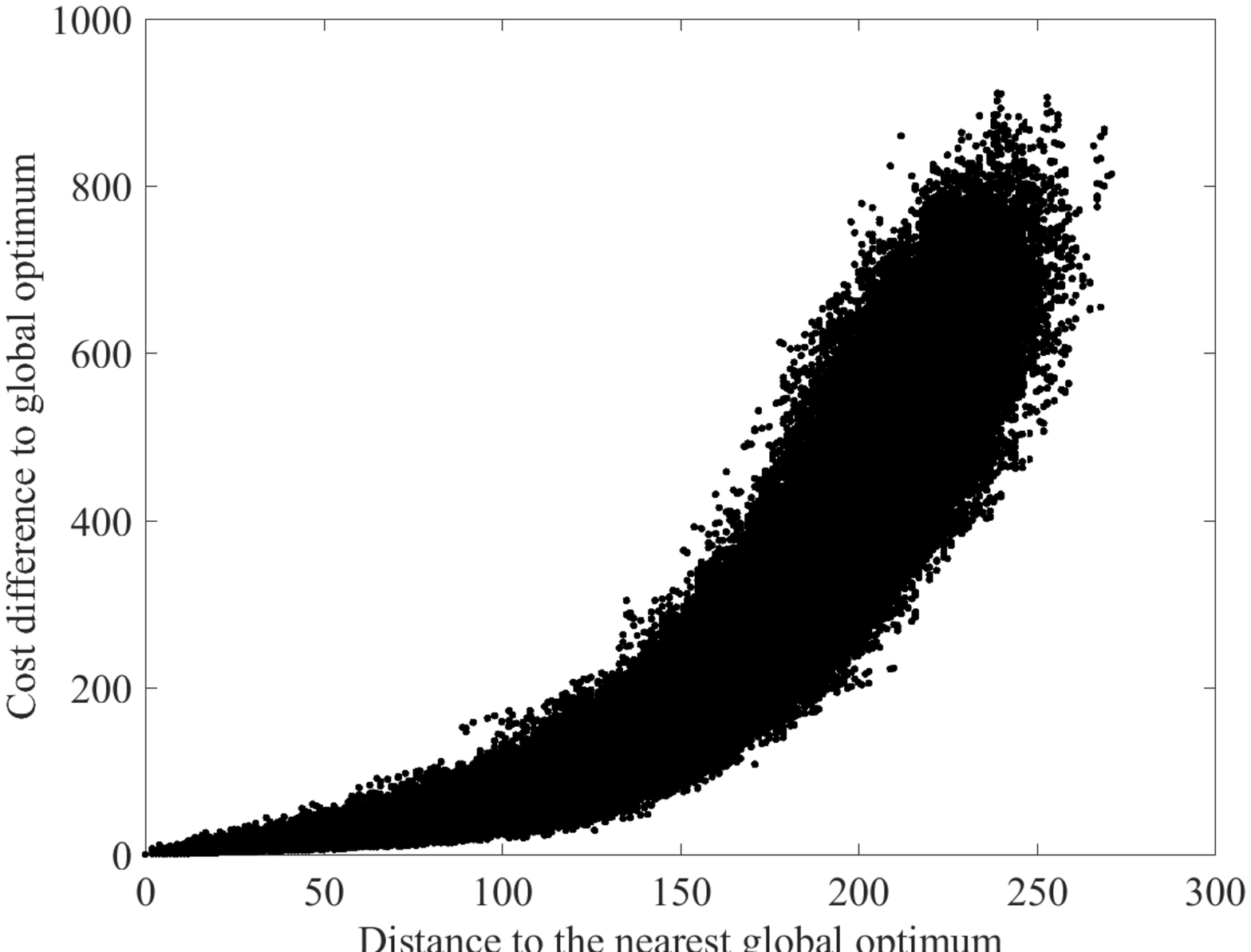}}
\hspace{0.01\textwidth}
  \subfigure[u1432]{
  \label{fig:cvd_u1432}
  \includegraphics[width=0.22\textwidth]{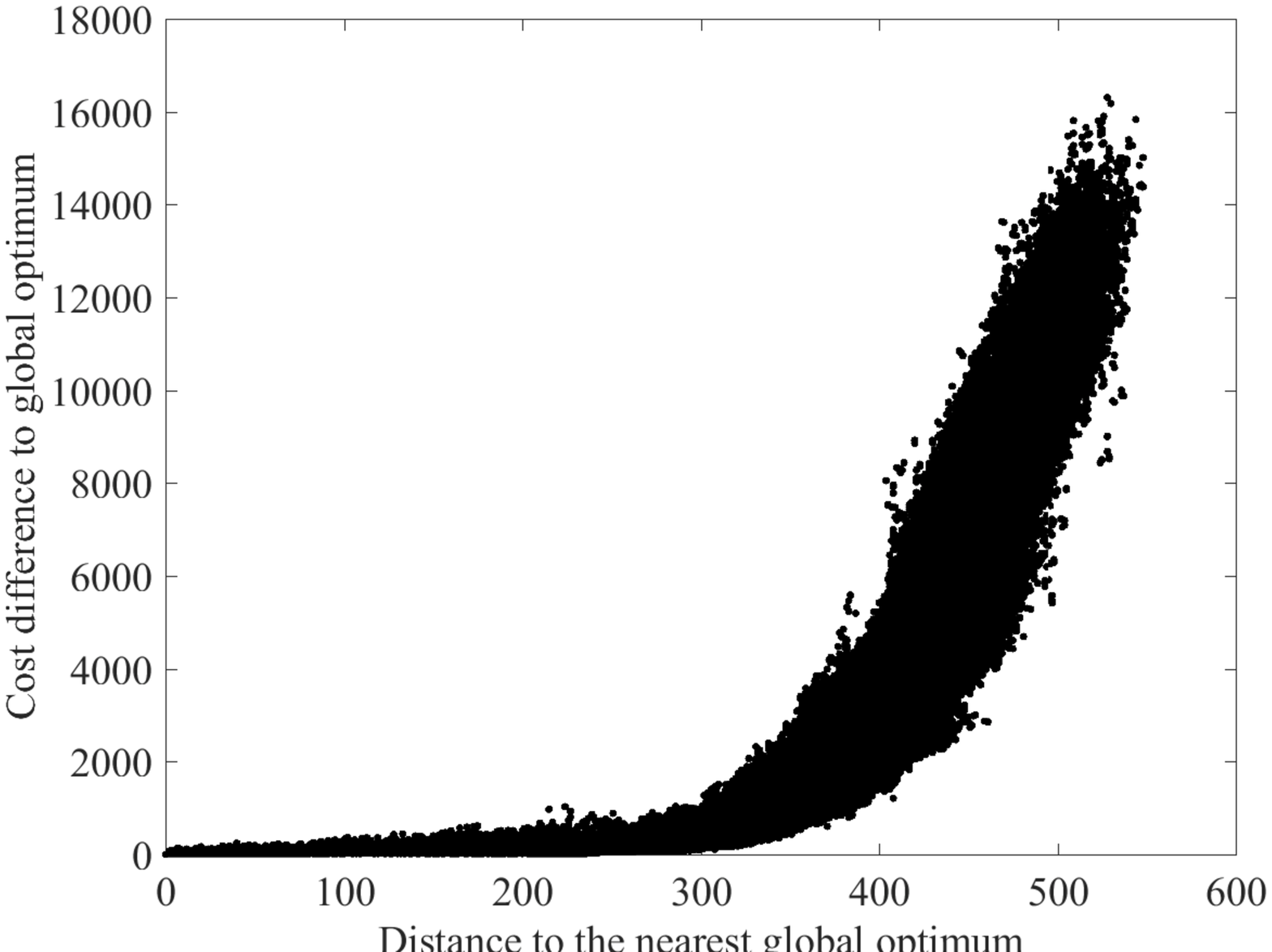}}
  \caption{The scatter plots of the recorded best solutions found so far during \text{1,000} runs of GLS and \text{1,000} runs of EB-GLS on eight selected instances. The cost difference to the globally optimal cost (vertical axis) is plotted against the distance to the nearest globally optimal solution (horizontal axis)}\label{fig:CvD_8_ins} 
\end{figure}

In Section \ref{sec:big_valley}, we state that att532 has the big valley structure, while u2319 does not have the big valley structure. Our statements are based on the landscape sampling experiment conducted on these two instances, in which \text{1,000} runs of GLS and \text{1,000} runs of EB-GLS are executed until finding the global optimum. During each run, the best solutions found so far are recorded and the final global optimum is also recorded. In fact, we conduct the same landscape sampling experiment on another eight instances. By analyzing the results, we conclude that all these eight instances satisfy the requirements of the big valley structure we defined in Section \ref{sec:big_valley}. Table~\ref{tbl:landscape_8_ins} shows the landscape sampling results on these eight instances. Fig. \ref{fig:CvD_8_ins} shows the scatter plots of the recorded best solutions found so far.

\begin{table}[H]
\caption{The landscape sampling results on eight selected TSPLIB instances. $N_{opt}$ is the number of the unique global optima found by these \text{2,000} runs. $D_{o,min}$ ($D_{o,avr}$,$D_{o,max}$) is the minimum (average, maximum) distance between the global optima. The last column is the FDC value of the recorded best solutions found so far during the \text{2,000} runs.}
\centering
\label{tbl:landscape_8_ins} 
\begin{tabular}{l|l|l|l|l|l|l}
  \hline
  Instances & City Num & $N_{opt}$ & $D_{o,min}$ & $D_{o,avr}$ & $D_{o,max}$ & FDC \\
  \hline
  rd400 & 400 & 8 & 3 & 21 & 36 & 0.83\\
  gr431 & 431 & 2 & 13 & 13 & 13 & 0.76\\
  pcb442 & 442 & 2000 & 6 & 40 & 69 & 0.79\\
  pa561 & 561 & 2000 & 6 & 44 & 82 & 0.83\\
  u574 & 574 & 4 & 2 & 4 & 6 & 0.84\\
  rat575 & 575 & 2 & 3 & 3 & 3 & 0.85\\
  rat783 & 783 & 811 & 2 & 15 & 31 & 0.89\\
  u1432 & 1432 & 2000 & 191 & 269 & 342 & 0.82\\
\hline
\end{tabular}
\end{table}
\end{appendices}

In Table \ref{tbl:cmprsn_rslt} we present the comparison results between EB-GLS and GLS on the 33 TSPLIB instances with more than 1000 cities. Table \ref{tbl:cmprsn_rslt_c1} shows the comparison results on the other 76 TSPLIB instances with less than 1000 cities.

\begin{table}
\caption{Comparison results between EB-GLS and GLS on the TSPLIB instances with less than 1000 cities, the better metric values are marked by bold texts}
\centering
\label{tbl:cmprsn_rslt_c1} 
\resizebox{\textwidth}{!}{
\begin{tabular}{l|c|c|c|r|r|c|r|r|c}
\hline
\multirow{2}{*}{Instance} & \multirow{2}{*}{\begin{minipage}{28pt}\tiny{Max Runtime (s)}\end{minipage}} & \multicolumn{2}{c|}{Success of 100} & \multicolumn{2}{c|}{Average Excess (\%)} & \multirow{2}{*}{\begin{minipage}{30pt}Excess P-value\end{minipage}} & \multicolumn{2}{c|}{Average Runtime (s)} & \multirow{2}{*}{\begin{minipage}{30pt}Runtime P-value\end{minipage}} \\ 
\cline{3-4}\cline{5-6}\cline{8-9}
&& \scriptsize{GLS} & \scriptsize{EB-GLS} & \multicolumn{1}{c|}{\scriptsize{GLS}} & \multicolumn{1}{c|}{\scriptsize{EB-GLS}} && \multicolumn{1}{c|}{\scriptsize{GLS}} & \multicolumn{1}{c|}{\scriptsize{EB-GLS}} & \\
\hline
burma14 & 2 & 100 & 100 & 0.0000 & 0.0000 $\approx$ & - & 0.0000 & 0.0000 $\approx$ & -  \\
ulysses16 & 2 & 100 & 100 & 0.0000 & 0.0000 $\approx$ & - & 0.0000 & 0.0000 $\approx$ & -  \\
gr17 & 2 & 100 & 100 & 0.0000 & 0.0000 $\approx$ & - & 0.0000 & 0.0000 $\approx$ & -  \\
gr21 & 3 & 100 & 100 & 0.0000 & 0.0000 $\approx$ & - & 0.0000 & 0.0000 $\approx$ & -  \\
ulysses22 & 3 & 100 & 100 & 0.0000 & 0.0000 $\approx$ & - & 0.0000 & 0.0000 $\approx$ & -  \\
gr24 & 3 & 100 & 100 & 0.0000 & 0.0000 $\approx$ & - & 0.0000 & 0.0000 $\approx$ & -  \\
fri26 & 3 & 100 & 100 & 0.0000 & 0.0000 $\approx$ & - & \textbf{0.0002} & 0.0006 $\approx$ & 1.51e-01  \\
bayg29 & 3 & 100 & 100 & 0.0000 & 0.0000 $\approx$ & - & \textbf{0.0006} & 0.0022 $-$ & 1.15e-03  \\
bays29 & 3 & 100 & 100 & 0.0000 & 0.0000 $\approx$ & - & 0.0009 & \textbf{0.0007} $\approx$ & 6.05e-01  \\
dantzig42 & 5 & 100 & 100 & 0.0000 & 0.0000 $\approx$ & - & \textbf{0.0006} & 0.0010 $\approx$ & 3.00e-01  \\
swiss42 & 5 & 100 & 100 & 0.0000 & 0.0000 $\approx$ & - & 0.0016 & \textbf{0.0006} $+$ & 2.43e-02  \\
att48 & 5 & 100 & 100 & 0.0000 & 0.0000 $\approx$ & - & 0.0048 & \textbf{0.0037} $\approx$ & 2.63e-01  \\
gr48 & 5 & 100 & 100 & 0.0000 & 0.0000 $\approx$ & - & 0.0067 & \textbf{0.0042} $\approx$ & 1.86e-01  \\
hk48 & 5 & 100 & 100 & 0.0000 & 0.0000 $\approx$ & - & \textbf{0.0017} & 0.0030 $-$ & 4.17e-02  \\
eil51 & 6 & 100 & 100 & 0.0000 & 0.0000 $\approx$ & - & \textbf{0.0103} & 0.0106 $\approx$ & 9.98e-01  \\
berlin52 & 6 & 100 & 100 & 0.0000 & 0.0000 $\approx$ & - & 0.0016 & \textbf{0.0010} $\approx$ & 2.09e-01  \\
brazil58 & 6 & 100 & 100 & 0.0000 & 0.0000 $\approx$ & - & 0.0032 & \textbf{0.0031} $\approx$ & 7.99e-01  \\
st70 & 7 & 100 & 100 & 0.0000 & 0.0000 $\approx$ & - & 0.0185 & \textbf{0.0119} $+$ & 5.81e-04  \\
eil76 & 8 & 100 & 100 & 0.0000 & 0.0000 $\approx$ & - & 0.0099 & \textbf{0.0082} $\approx$ & 9.03e-02  \\
pr76 & 8 & 100 & 100 & 0.0000 & 0.0000 $\approx$ & - & 0.0407 & \textbf{0.0148} $+$ & 8.73e-05  \\
gr96 & 10 & 100 & 100 & 0.0000 & 0.0000 $\approx$ & - & \textbf{0.0262} & 0.0267 $\approx$ & 5.56e-01  \\
rat99 & 10 & 100 & 100 & 0.0000 & 0.0000 $\approx$ & - & 0.0325 & \textbf{0.0191} $+$ & 9.54e-07  \\
kroA100 & 10 & 100 & 100 & 0.0000 & 0.0000 $\approx$ & - & 0.0158 & \textbf{0.0103} $+$ & 4.46e-05  \\
kroB100 & 10 & 100 & 100 & 0.0000 & 0.0000 $\approx$ & - & 0.0458 & \textbf{0.0280} $+$ & 7.80e-03  \\
kroC100 & 10 & 100 & 100 & 0.0000 & 0.0000 $\approx$ & - & 0.0247 & \textbf{0.0069} $+$ & 2.55e-19  \\
kroD100 & 10 & 100 & 100 & 0.0000 & 0.0000 $\approx$ & - & 0.0287 & \textbf{0.0133} $+$ & 4.78e-12  \\
kroE100 & 10 & 100 & 100 & 0.0000 & 0.0000 $\approx$ & - & 0.0534 & \textbf{0.0314} $+$ & 2.02e-05  \\
rd100 & 10 & 100 & 100 & 0.0000 & 0.0000 $\approx$ & - & 0.0354 & \textbf{0.0206} $+$ & 4.10e-07  \\
eil101 & 11 & 100 & 100 & 0.0000 & 0.0000 $\approx$ & - & \textbf{0.0195} & 0.0204 $\approx$ & 9.70e-01  \\
lin105 & 11 & 100 & 100 & 0.0000 & 0.0000 $\approx$ & - & 0.0202 & \textbf{0.0125} $+$ & 3.36e-06  \\
pr107 & 11 & 100 & 100 & 0.0000 & 0.0000 $\approx$ & - & 0.7517 & \textbf{0.1438} $+$ & 2.35e-23  \\
gr120 & 12 & 100 & 100 & 0.0000 & 0.0000 $\approx$ & - & 0.1438 & \textbf{0.0493} $+$ & 2.36e-15  \\
pr124 & 13 & 100 & 100 & 0.0000 & 0.0000 $\approx$ & - & \textbf{0.0246} & 0.0337 $-$ & 3.15e-04  \\
bier127 & 13 & 100 & 100 & 0.0000 & 0.0000 $\approx$ & - & 0.3955 & \textbf{0.1055} $+$ & 7.47e-21  \\
ch130 & 13 & 100 & 100 & 0.0000 & 0.0000 $\approx$ & - & 0.0895 & \textbf{0.0642} $+$ & 1.67e-05  \\
pr136 & 14 & 99 & \textbf{100} & 0.0001 & \textbf{0.0000} $\approx$ & 3.22e-01 & 1.7463 & \textbf{0.1631} $+$ & 3.44e-18  \\
gr137 & 14 & 100 & 100 & 0.0000 & 0.0000 $\approx$ & - & 0.0907 & \textbf{0.0429} $+$ & 2.01e-17  \\
pr144 & 15 & 100 & 100 & 0.0000 & 0.0000 $\approx$ & - & \textbf{0.0811} & 0.1134 $-$ & 3.58e-03  \\
ch150 & 15 & 100 & 100 & 0.0000 & 0.0000 $\approx$ & - & 0.2003 & \textbf{0.1230} $+$ & 4.34e-07  \\
kroA150 & 15 & 100 & 100 & 0.0000 & 0.0000 $\approx$ & - & 0.3707 & \textbf{0.0848} $+$ & 9.34e-14  \\
kroB150 & 15 & 100 & 100 & 0.0000 & 0.0000 $\approx$ & - & 0.3672 & \textbf{0.2334} $+$ & 1.12e-06  \\
pr152 & 16 & 100 & 100 & 0.0000 & 0.0000 $\approx$ & - & 1.8919 & \textbf{0.6639} $+$ & 1.68e-09  \\
u159 & 16 & 100 & 100 & 0.0000 & 0.0000 $\approx$ & - & 0.0658 & \textbf{0.0628} $\approx$ & 9.21e-01  \\
si175 & 18 & 100 & 100 & 0.0000 & 0.0000 $\approx$ & - & 5.4633 & \textbf{1.8940} $+$ & 1.65e-20  \\
brg180 & 18 & 100 & 100 & 0.0000 & 0.0000 $\approx$ & - & \textbf{0.0044} & 0.0049 $\approx$ & 8.02e-01  \\
rat195 & 20 & 100 & 100 & 0.0000 & 0.0000 $\approx$ & - & 0.4984 & \textbf{0.2696} $+$ & 2.83e-05  \\
d198 & 20 & 75 & \textbf{96} & 0.0016 & \textbf{0.0004} $+$ & 3.28e-05 & 14.7688 & \textbf{4.7853} $+$ & 2.51e-22  \\
kroA200 & 20 & 100 & 100 & 0.0000 & 0.0000 $\approx$ & - & 1.7318 & \textbf{0.1695} $+$ & 1.26e-32  \\
kroB200 & 20 & 100 & 100 & 0.0000 & 0.0000 $\approx$ & - & 1.2219 & \textbf{0.0886} $+$ & 5.37e-32  \\
gr202 & 21 & 100 & 100 & 0.0000 & 0.0000 $\approx$ & - & 1.8843 & \textbf{0.6493} $+$ & 6.34e-19  \\
ts225 & 23 & 100 & 100 & 0.0000 & 0.0000 $\approx$ & - & 0.7552 & \textbf{0.4194} $+$ & 4.46e-05  \\
tsp225 & 23 & 99 & \textbf{100} & 0.0008 & \textbf{0.0000} $\approx$ & 3.22e-01 & 2.9281 & \textbf{0.6930} $+$ & 2.74e-15  \\
pr226 & 23 & 93 & \textbf{98} & 0.0008 & \textbf{0.0006} $\approx$ & 9.29e-02 & 6.5625 & \textbf{3.0864} $+$ & 1.89e-08  \\
gr229 & 23 & 89 & \textbf{100} & 0.0010 & \textbf{0.0000} $+$ & 6.76e-04 & 8.0847 & \textbf{1.2958} $+$ & 1.25e-22  \\
gil262 & 27 & 100 & 100 & 0.0000 & 0.0000 $\approx$ & - & 2.3682 & \textbf{1.0298} $+$ & 3.09e-10  \\
pr264 & 27 & 100 & 100 & 0.0000 & 0.0000 $\approx$ & - & 0.9685 & \textbf{0.4527} $+$ & 1.68e-15  \\
a280 & 28 & 100 & 100 & 0.0000 & 0.0000 $\approx$ & - & 0.1767 & \textbf{0.1731} $\approx$ & 3.19e-01  \\
pr299 & 30 & 94 & \textbf{100} & 0.0001 & \textbf{0.0000} $+$ & 1.33e-02 & 9.7268 & \textbf{0.4571} $+$ & 1.12e-31  \\
lin318 & 32 & 100 & 100 & 0.0000 & 0.0000 $\approx$ & - & 2.3201 & \textbf{1.2127} $+$ & 8.33e-09  \\
rd400 & 40 & 87 & \textbf{93} & 0.0009 & \textbf{0.0005} $\approx$ & 1.59e-01 & 20.4316 & \textbf{6.8556} $+$ & 2.67e-19  \\
fl417 & 42 & 1 & \textbf{83} & 0.0333 & \textbf{0.0123} $+$ & 1.05e-18 & 41.7864 & \textbf{18.7175} $+$ & 2.85e-29  \\
gr431 & 44 & 4 & \textbf{71} & 0.0072 & \textbf{0.0005} $+$ & 3.25e-28 & 43.1186 & \textbf{22.0183} $+$ & 9.38e-22  \\
pr439 & 44 & 1 & \textbf{58} & 0.0598 & \textbf{0.0140} $+$ & 3.97e-26 & 43.8654 & \textbf{26.0844} $+$ & 3.09e-18  \\
pcb442 & 45 & 28 & \textbf{94} & 0.0087 & \textbf{0.0014} $+$ & 1.22e-18 & 37.2381 & \textbf{8.9825} $+$ & 2.66e-24  \\
d493 & 50 & 1 & \textbf{15} & 0.0135 & \textbf{0.0065} $+$ & 2.10e-20 & 49.9264 & \textbf{47.3198} $+$ & 2.66e-04  \\
att532 & 54 & 6 & \textbf{98} & 0.0229 & \textbf{0.0012} $+$ & 1.33e-32 & 52.0478 & \textbf{10.4495} $+$ & 1.17e-35  \\
ali535 & 54 & 0 & \textbf{96} & 0.0337 & \textbf{0.0010} $+$ & 2.81e-36 & 54.0000 & \textbf{16.4560} $+$ & 1.12e-36  \\
si535 & 54 & 0 & \textbf{2} & 0.2140 & \textbf{0.0387} $+$ & 1.77e-33 & 54.0000 & \textbf{53.4750} $\approx$ & 1.58e-01  \\
pa561 & 57 & 69 & \textbf{99} & 0.0119 & \textbf{0.0004} $+$ & 7.99e-09 & 33.7124 & \textbf{10.0504} $+$ & 7.84e-20  \\
u574 & 58 & 31 & \textbf{100} & 0.0076 & \textbf{0.0000} $+$ & 2.44e-23 & 50.9637 & \textbf{7.4151} $+$ & 3.01e-34  \\
rat575 & 58 & 6 & \textbf{53} & 0.0337 & \textbf{0.0080} $+$ & 4.89e-24 & 57.1423 & \textbf{38.1065} $+$ & 8.13e-14  \\
p654 & 66 & 0 & \textbf{5} & 0.1657 & \textbf{0.0461} $+$ & 1.30e-31 & 66.0000 & \textbf{65.1685} $+$ & 2.42e-02  \\
d657 & 66 & 0 & 0 & 0.0258 & \textbf{0.0038} $+$ & 1.68e-31 & 66.0000 & 66.0000 $\approx$ & -  \\
gr666 & 67 & 0 & \textbf{25} & 0.0417 & \textbf{0.0158} $+$ & 1.81e-16 & 67.0000 & \textbf{59.0446} $+$ & 1.07e-07  \\
u724 & 73 & 2 & \textbf{58} & 0.0428 & \textbf{0.0069} $+$ & 3.68e-26 & 72.5028 & \textbf{50.4334} $+$ & 1.59e-17  \\
rat783 & 79 & 24 & \textbf{100} & 0.0151 & \textbf{0.0000} $+$ & 1.07e-26 & 70.4217 & \textbf{11.3429} $+$ & 6.42e-35  \\
\hline
\end{tabular}
}
\end{table}

\end{document}